\pdfoutput=1

\documentclass[sigconf,nonacm]{acmart}

\usepackage{booktabs}
\usepackage{array}
\usepackage{tabularx}
\usepackage{pifont}
\usepackage{xspace}
\usepackage{tikz}
\usetikzlibrary{positioning,arrows.meta,fit,backgrounds,calc}
\usepackage[most]{tcolorbox}
\usepackage{listings}

\tcbset{
  promptbox/.style={enhanced, breakable, colback=gray!4, colframe=black!25,
    boxrule=0.5pt, arc=2mm, left=1.5mm, right=1.5mm, top=1mm, bottom=1mm, boxsep=1mm}
}
\lstdefinestyle{promptstyle}{
  basicstyle=\ttfamily\scriptsize, columns=fullflexible, keepspaces=true,
  breaklines=true, breakatwhitespace=false, showstringspaces=false,
  aboveskip=0pt, belowskip=0pt
}

\newcommand{\method}{\textsc{Agents-as-Peers}\xspace}
\newcommand{\cmark}{\ding{51}}
\newcommand{\xmark}{\ding{55}}
\newcommand{\pmark}{$\triangle$}
\newcommand{\truthfulQAPeerPearson}{0.904}

\title[\method: Multi-Agent Peer Assessment for LLM Evaluation]{\method: Bias-Controlled LLM Evaluation\texorpdfstring{\\}{ }as Multi-Agent Peer Assessment}

\author{Yanki Margalit}
\affiliation{
  \institution{Caura.ai}
  \country{Israel}}

\author{Erni Avram}
\affiliation{
  \institution{Caura.ai}
  \country{Israel}}

\author{Ran Taig}
\affiliation{
  \institution{Caura.ai}
  \country{Israel}}

\author{Seffi Cohen}
\affiliation{
  \department{Computer Science and Information}
  \institution{Ben-Gurion University of the Negev}
  \country{Israel}}

\author{Oded Margalit}
\affiliation{
  \department{Computer Science and Information}
  \institution{Ben-Gurion University of the Negev}
  \country{Israel}}

\author{Nurit Cohen-Inger}
\affiliation{
  \department{Computer Science and Information}
  \institution{Ben-Gurion University of the Negev}
  \country{Israel}}

\begin{abstract}
Large language model (LLM) evaluation commonly relies on fixed human-authored benchmarks or centralized human/model judges. Fixed benchmarks can age and enter model training corpora, while centralized judges introduce their own scoring preferences. We cast LLM evaluation as a multi-agent peer-assessment problem and introduce \method, a distributed evaluation protocol in which every agent generates tasks, answers the shared task set, and judges the answers of its peers, with no privileged evaluator. Self-ratings are excluded from the ranking, and judging is repeated under controlled identity and presentation regimes so that self bias, identity effects, presentation-position sensitivity, and judge severity are measured rather than left implicit. In an instantiation with 12 LLMs and 420 agent-generated questions, \method aggregates $55{,}440$ non-self peer judgments per regime. Most agents assign their own answers higher scores than those answers receive from peer judges; visible identity changes scores for some agents; the first fixed answer slot is associated with scores $0.39$ points higher than the shuffled baseline; and judges differ markedly in severity. On labeled benchmarks, peer scores align strongly with accuracy on TruthfulQA (Pearson $r=\truthfulQAPeerPearson$) and GSM8K ($r=0.873$). On TruthfulQA, the peer-score correlation is substantially higher than the self-score correlation ($r=0.538$). A benchmark-overlap screen and item-level human audit further characterize the endogenous task set. These results support \method as a cohort-relative measurement protocol whose ranking is reported together with the biases and rater effects that shape it.
Code and dataset are available at: \url{https://github.com/caura-ai/caura-PeerRank}.
\end{abstract}


\begin{document}


\maketitle

\begin{figure*}[t]
\centering
\begin{tikzpicture}[
  font=\sffamily\footnotesize,
  >={Stealth[length=2mm]},
  phase/.style={draw=#1!65!black, fill=#1!7, rounded corners=3pt, line width=0.6pt,
                text width=3.72cm, minimum height=3.05cm, align=left, inner sep=4pt, anchor=north west},
  ptitle/.style={font=\sffamily\bfseries\footnotesize},
  agent/.style={circle, draw=black!55, fill=white, minimum size=6.2mm, inner sep=0pt, font=\sffamily\scriptsize},
  role/.style={font=\sffamily\scriptsize\itshape, text=black!70},
  regime/.style={draw=black!45, fill=white, rounded corners=2pt, text width=3.72cm, align=left,
                 inner sep=3pt, minimum height=1.2cm, anchor=north west, font=\sffamily\scriptsize},
  flow/.style={->, line width=0.8pt, draw=black!65},
  thin/.style={->, line width=0.5pt, draw=black!45, dashed}
]
\node[agent] (m1) at (0.35,0.55) {$m_1$};
\node[agent, right=1.5mm of m1] (m2) {$m_2$};
\node[right=0.5mm of m2, font=\sffamily\small] (dots) {$\cdots$};
\node[agent, right=0.5mm of dots] (mK) {$m_K$};
\node[anchor=west, align=left, font=\sffamily\footnotesize] at ($(mK.east)+(0.2,0)$)
  {\textbf{Cohort of $K$ LLM agents.} Every agent plays every role; no agent holds privileged evaluative authority.};
\node[phase=blue]   (p1) at (0,-0.35) {\textbf{1\;\;Generate tasks}\\[2pt]
  Each agent writes $n$ questions in five categories.\\[2pt]
  Corpus $\mathcal{Q}$, $|\mathcal{Q}|=Kn$ (here $12\times35=420$), used as generated: no filtering or editing.};
\node[phase=teal]   (p2) at (4.48,-0.35) {\textbf{2\;\;Answer}\\[2pt]
  Every agent answers every question.\\[2pt]
  One shared retrieval provider for all agents, used only for current-events items.\\[2pt]
  Answers $a_{j,q}$.};
\node[phase=orange] (p3) at (8.96,-0.35) {\textbf{3\;\;Peer judging}\\[2pt]
  Every agent scores every answer on a 1--10 rubric, without web access.\\[2pt]
  Score tensor $\mathbf{S}=\{s_{i,j,q}\}$\\
  (judge $\times$ respondent $\times$ question), repeated under three regimes.};
\node[phase=violet] (p4) at (13.44,-0.35) {\textbf{4\;\;Aggregate \& report}\\[2pt]
  Peer score $P_j$, self-ratings excluded $\rightarrow$ ranking.\\[2pt]
  Report self bias, identity effects, and presentation-position sensitivity.\\[2pt]
  Judge severity $G_i$, agreement, Elo check.};
\draw[flow] (p1.east) -- (p2.west);
\draw[flow] (p2.east) -- (p3.west);
\draw[flow] (p3.east) -- (p4.west);
\node[role, anchor=south] at ($(p1.north)+(0.9,0.02)$) {agents as generators};
\node[role, anchor=south] at ($(p2.north)+(0.9,0.02)$) {agents as respondents};
\node[role, anchor=south] at ($(p3.north)+(0.9,0.02)$) {agents as judges};
\node[regime] (r1) at (4.48,-3.72) {\textbf{Shuffle+blind} (primary)\\order randomized, identities hidden\\$\Rightarrow$ ranking $P_j$ and self bias\\$\Delta^{\mathrm{self}}=$ self score $-\,P_j$};
\node[regime] (r2) at (8.96,-3.72) {\textbf{Shuffle-only}\\order randomized, identities visible\\$\Rightarrow$ identity effects\\$\Delta^{\mathrm{id}}=P^{\mathrm{shuffle}}_j-P_j$};
\node[regime] (r3) at (13.44,-3.72) {\textbf{Blind-only}\\order fixed, identities hidden\\$\Rightarrow$ presentation-position sensitivity\\$\Delta^{\mathrm{pres}}=P^{\mathrm{blind}}_j-P_j$};
\coordinate (bus) at ($(p3.south)+(0,-0.16)$);
\draw[line width=0.5pt, draw=black!45] (p3.south) -- (bus);
\draw[line width=0.5pt, draw=black!45] (bus -| r1.north) -- (bus -| r3.north);
\foreach \r in {r1,r2,r3} {\draw[->, line width=0.5pt, draw=black!45] (bus -| \r.north) -- (\r.north);}
\node[anchor=north west, text width=3.72cm, align=left, font=\sffamily\scriptsize, draw=black!35, dashed, rounded corners=2pt, inner sep=3pt, minimum height=1.2cm] (val) at (0,-3.72)
  {\textbf{External check} (outside the protocol): run the same judging on labeled benchmarks (TruthfulQA, GSM8K) and correlate $P_j$ with accuracy.};
\end{tikzpicture}
\caption{The \method protocol. A cohort of LLM agents acts symmetrically as question generators, respondents, and judges. Web retrieval is available only when answering current-events questions and is disabled during judging. Every judgment is collected under three regimes; the shuffle+blind regime yields the ranking, and contrasts with the other two regimes yield the identity effects and presentation-position sensitivity reported alongside it.}
\label{fig:peerrank-pipeline}
\Description{Conceptual pipeline of Agents-as-Peers: a cohort of LLM agents generates questions, answers every question, evaluates peer answers under bias-control regimes, and aggregates peer scores; an external benchmark check is shown as a separate validation step.}
\end{figure*}

\section{Introduction}

Large language models (LLMs) are increasingly deployed as agents that retrieve information, use tools, and interact with other models. Evaluating these systems requires both suitable tasks and reliable judgments. Fixed human-authored benchmarks provide comparability, but can become stale or enter training data~\cite{inger2025forget,chen2025benchmarking,balloccu-etal-2024-leak}; model judges make evaluation scalable, but introduce position, identity, and self-preference effects~\cite{shi2025judging,self_preference_bias_2024,zhao2024justice}. These limitations motivate distributing task creation and judgment across a cohort of models. The central question is whether such a cohort can produce informative evaluations when its members both assess one another and define the tasks on which they are assessed.

This setting connects LLM evaluation to established problems in multi-agent systems. Peer grading addresses differences in evaluator reliability and severity~\cite{piech2013tuned,wright2015mechanical}; impartial peer selection studies how agents can assess competitors without controlling their own selection~\cite{holzman2013impartial,alon2011sum,kurokawa2015impartial,aziz2019strategyproof}; and peer prediction considers informative reporting without observable ground truth~\cite{miller2005eliciting}. For LLM peer assessment, the challenge is to determine whether collective scores track correctness and how they depend on the participating judges and the tasks they generate.

We study three questions. \emph{First}, when objective labels are available, do scores produced by an LLM peer cohort align with correctness? \emph{Second}, how strongly are those scores affected by self-evaluation, visible model identity, answer presentation, and judge-specific severity? \emph{Third}, what task distribution emerges when the evaluated agents themselves generate the evaluation items? To study these questions, we introduce \method, a protocol in which every agent acts symmetrically as \emph{question generator}, \emph{respondent}, and \emph{judge} (Figure~\ref{fig:peerrank-pipeline}). Current-events questions use one retrieval layer shared across agents; judging is performed without web access; and judgments are collected under three controlled regimes that manipulate answer order and model identity. Rankings exclude self-ratings, while contrasts between regimes expose identity and presentation sensitivity instead of absorbing them into a single score.

\method is a cohort-relative measurement protocol, not an absolute correctness benchmark. We therefore validate its scoring signal on labeled benchmarks, audit the generated task set, and report bias and rater diagnostics alongside the ranking.

\subsection*{Our contributions are threefold:}
\begin{enumerate}
    \item \textbf{A symmetric peer-assessment protocol for LLM evaluation.} The same cohort generates tasks, answers them, and judges the responses. Human designers specify the categories, prompts, and rubric; models supply the evaluation items and judgments. Rankings aggregate peer scores while excluding self-ratings.
    \item \textbf{An evaluation of how judging conditions shape peer scores.} Controlled identity and answer-order regimes expose scoring sensitivity, while self--peer comparisons and judge-calibration analyses characterize evaluator differences. The protocol reports these diagnostics alongside the ranking.
    \item \textbf{Evidence on both scoring validity and task quality.} In a study of 12 LLMs and 420 generated questions, we test peer scores against labeled benchmark accuracy and examine ranking sensitivity to aggregation choices. A benchmark-overlap screen and human audit reveal limited direct benchmark reproduction but substantial internal task redundancy, showing why generated tasks require scrutiny alongside their judges.
\end{enumerate}

\section{Related Work}
\label{sec:related}

\paragraph{Peer assessment and peer selection in multi-agent systems.}
Peer-based evaluation predates LLMs. Calibrated peer grading models grader bias and reliability when aggregating noisy student grades~\cite{piech2013tuned}, and Mechanical TA combines peer and expert grading for high-stakes assessment~\cite{wright2015mechanical}. Peer-selection work focuses on \emph{impartiality}: an agent's report should not directly determine its own selection outcome~\cite{holzman2013impartial,alon2011sum,kurokawa2015impartial,aziz2019strategyproof}, while peer prediction addresses truthful reporting without observable ground truth~\cite{miller2005eliciting}. \method draws on this broader peer-evaluation perspective but studies systematic scoring biases in prompted LLM evaluators rather than strategic misreporting; incentive-aware extensions are discussed in Section~\ref{sec:discussion}.

\paragraph{Rater models and item analysis.}
Educational measurement similarly separates item properties from rater effects: item-response analysis characterizes difficulty and coverage~\cite{lord1980applications}, while many-facet Rasch models estimate rater severity and consistency~\cite{eckes2023introduction}. We use this distinction to analyze judge agreement, generosity, and the endogenous task distribution.

\paragraph{LLM-as-a-judge and LLM peer review.}
LLM judges can correlate strongly with human preferences~\cite{zheng2023judging_llm_as_judge} and support reference-free evaluation~\cite{liu2023geval}; evaluator ensembles and debates distribute judgment across multiple models~\cite{chan2024chateval}. Chatbot Arena aggregates pairwise human preferences into Elo-style ratings~\cite{chiang2024chatbot_arena,elo1978rating}. More recent peer-review and label-free systems, including PRE~\cite{chu2024pre}, UPME~\cite{zhang2025upme}, and AutoBench~\cite{loi2025autobench}, cast models as assessors, task generators, or contestants. \method differs in combining three properties in one protocol: the cohort generates the task distribution, every cohort member participates as a judge, and bias controls are measured and reported as part of the evaluation output.

\paragraph{Bias and reliability of automated evaluation.}
LLM judges exhibit position bias~\cite{shi2025judging}, self-preference~\cite{self_preference_bias_2024}, and identity or presentation sensitivity~\cite{zhao2024justice,thakur2025judging}. Work on judge reliability and correctness~\cite{wei2025iclrw-systematic,lambert2025rewardbench} further motivates randomized order, identity blinding, self-rating exclusion, and explicit rater diagnostics.

\paragraph{Benchmarks and open-world agent evaluation.}
Static benchmarks such as MMLU, SQuAD, and GSM8K~\cite{hendrycks2021measuring,rajpurkar2016squad,cobbe2021gsm8k} and holistic suites such as HELM~\cite{liang2022helm} are repeatable, but they can lag behind model capabilities~\cite{wang2024mmlu}, become stale~\cite{inger2025forget}, incur contamination and maintenance costs~\cite{chen2025benchmarking,balloccu-etal-2024-leak}, and drift from real-world use~\cite{rontogiannis2025interactive,you2024llmevolve}. Agent benchmarks such as AgentBench~\cite{liu2023agentbench} and WebArena~\cite{zhou2023webarena} improve interaction realism, while multi-agent approaches can generate evolving evaluation tasks~\cite{wang-etal-2025-benchmark}. \method combines endogenous task generation with controlled peer judging and live retrieval for current-events items. Table~\ref{tab:related-work-summary} summarizes the relevant design axes.

\begin{table}[t]
\centering
\caption{Comparison of evaluation paradigms. \emph{Gen. tasks}: tasks are generated at evaluation time; \emph{Auto. score}: scoring requires no human judgment at evaluation time; \emph{Live}: evaluation can access a live environment or retrieval source; \emph{Agents judge}: evaluated agents participate as judges; \emph{Bias ctrl.}: explicit controls or analyses for evaluator bias. \pmark: partial or system-dependent support.}
\label{tab:related-work-summary}
\begin{tabularx}{\columnwidth}{>{\raggedright\arraybackslash}p{0.34\columnwidth} *{5}{>{\centering\arraybackslash}X}}
\toprule
\textbf{Approach} & \textbf{Gen. tasks} & \textbf{Auto. score} & \textbf{Live} & \textbf{Agents judge} & \textbf{Bias ctrl.} \\
\midrule
Static benchmarks                          & \xmark & \cmark & \xmark & \xmark & \xmark \\
Holistic suites (HELM)                     & \xmark & \cmark & \xmark & \xmark & \pmark \\
Agent benchmarks (AgentBench, WebArena)    & \xmark & \cmark & \cmark & \xmark & \xmark \\
LLM-as-a-judge (MT-Bench)                  & \xmark & \cmark & \xmark & \xmark & \pmark \\
Preference platforms (Chatbot Arena)       & \xmark & \xmark & \xmark & \xmark & \pmark \\
Human peer assessment and selection        & \xmark & \xmark & \xmark & \cmark & \pmark \\
LLM peer review (PRE, UPME, AutoBench)     & \pmark & \cmark & \xmark & \cmark & \pmark \\
\addlinespace
\textbf{\method (ours)}                  & \cmark & \cmark & \cmark & \cmark & \cmark \\
\bottomrule
\end{tabularx}
\end{table}

\section{Method}
\label{sec:protocol}

\paragraph{Setting and notation.}
Let $\mathcal{M}=\{m_1,\dots,m_K\}$ be a cohort of $K$ LLM agents. Every agent participates in three roles: question generator, respondent, and judge. Under dense judging, each question induces a directed peer-evaluation graph over the cohort: every agent evaluates every other agent, while self-edges are excluded from the ranking. Agent $m_i$ generates an indexed collection $\mathcal{Q}_i=(q_{i,1},\ldots,q_{i,n})$ of $n$ question instances. The evaluation corpus $\mathcal{Q}$ is the concatenation of these indexed collections, so identical strings generated by different agents remain distinct instances and $N=|\mathcal{Q}|=Kn$. For $q\in\mathcal{Q}$, let $a_{j,q}$ denote agent $m_j$'s answer and $s_{i,j,q}\in\{1,\dots,10\}$ the score assigned by judge $m_i$.

\paragraph{Design principles.}
The protocol follows four principles. \emph{Role symmetry}: every agent contributes equally many task instances and participates as both respondent and judge. \emph{Self-rating exclusion}: an agent's self-rating is never used in its peer score. \emph{Controlled information conditions}: agents answer under the same retrieval policy, while judging regimes vary only the intended identity and presentation manipulations. \emph{Measured rather than assumed neutrality}: self, identity, presentation, and rater effects are estimated explicitly instead of being folded into a single ranking.

\subsection*{Phase 1: Endogenous Task Generation}
Each agent independently generates $n$ questions over a fixed set of categories. We preserve all generated instances, including exact duplicates and paraphrases, without filtering, deduplication, or human editing. This keeps the task distribution endogenous to the cohort, but it also makes redundancy and benchmark reproduction possible; Section~\ref{sec:corpus} therefore audits the corpus rather than assuming it is novel or independent.

\subsection*{Phase 2: Web-Grounded Answering}
Every agent answers every question instance, $a_{j,q}=f_{m_j}(q,\mathcal{W})$, where $\mathcal{W}$ denotes an external search environment. Retrieval is provided through one shared layer: within a run, every agent uses the same grounding provider (e.g., Tavily or SerpAPI), and provider-native browsing tools are disabled. Retrieved snippets are included in the answering context but are not shown separately to judges. Retrieval is used only for current-events questions; all other categories are answered without retrieved evidence. Response time is recorded but does not affect scoring.

\subsection*{Phase 3: Peer Judging Under Controlled Regimes}
Every agent scores every submitted answer under a standardized rubric that prioritizes correctness and faithfulness, followed by completeness, clarity, and usefulness (full prompt in Appendix~\ref{app:prompts}). Judges assign an integer score from 1 to 10. Scores of 7--9 indicate mostly correct answers with minor omissions or imprecision; 4--6 indicate mixed or partial correctness, unclear reasoning, or missed constraints; and 1--3 indicate mostly incorrect, misleading, evasive, or unsupported answers. For subjective or creative questions, judges score instruction-following, coherence, and usefulness rather than factual correctness. Each judgment also includes an 8--20 word justification and flags from a fixed vocabulary. Web access is disabled during judging. Self-evaluations are collected for diagnostic purposes but excluded from peer-score aggregation.

To probe known identity and presentation effects~\cite{shi2025judging,self_preference_bias_2024}, the judging pass is repeated under three regimes: (i) \emph{shuffle-only}, with randomized answer order and visible model identities; (ii) \emph{blind-only}, with fixed answer order and hidden identities; and (iii) \emph{shuffle+blind}, with randomized order and hidden identities. Shuffle+blind is the primary reference condition because it removes the two directly manipulated cues. The blind-only contrast is interpreted as \emph{presentation-position sensitivity}, not as a fully isolated causal estimate of position bias, because fixed slots can remain correlated with respondent-specific characteristics.

\subsection*{Phase 4: Aggregation, Bias, and Judge Statistics}
Let $s^{\mathrm{SB}}$, $s^{\mathrm{S}}$, and $s^{\mathrm{B}}$ denote judgments from shuffle+blind, shuffle-only, and blind-only, respectively. The primary \emph{peer score} excludes self-ratings:
\begin{equation}
P_j=P^{\mathrm{SB}}_j=\mathbb{E}_{i\neq j,\,q}\left[s^{\mathrm{SB}}_{i,j,q}\right].
\end{equation}
Judge \emph{generosity}, the inverse of rater severity, is $G_i=\mathbb{E}_{j\neq i,\,q}[s^{\mathrm{SB}}_{i,j,q}]$. We report three score contrasts:
\begin{equation}
\Delta^{\mathrm{self}}_j=\mathbb{E}_{q}[s^{\mathrm{SB}}_{j,j,q}]-P_j,\quad
\Delta^{\mathrm{id}}_j=P^{\mathrm{S}}_j-P_j,\quad
\Delta^{\mathrm{pres}}_j=P^{\mathrm{B}}_j-P_j.
\end{equation}
Positive values indicate higher scores under self-evaluation, visible identity, or fixed presentation, respectively. Rankings sort agents by $P_j$. Raw judgment dispersion is summarized by $\sigma^2_j=\mathrm{Var}_{i\neq j,\,q}(s^{\mathrm{SB}}_{i,j,q})$; this quantity mixes item and judge variation and is therefore used as a descriptive dispersion measure rather than an uncertainty estimate.

Blinding removes explicit model names but may leave stylistic cues that allow judges to recognize their own outputs. We therefore interpret $\Delta^{\mathrm{self}}_j$ as a descriptive self--peer scoring gap. Self-recognition, stylistic preferences, and differences in judge severity are not separately identified by this contrast.

\paragraph{Aggregation sensitivity: Elo.}
The available within-judge, within-question comparisons comprise $254{,}195$ non-self answer pairs: $164{,}203$ with unequal scores and $89{,}992$ ties. We report sequential Elo ratings~\cite{elo1978rating} with $K_{\mathrm{Elo}}=32$ as an order-sensitive aggregation check. Because sequential Elo can depend on comparison order, we use it only to test whether the broad ordering is compatible with a pairwise aggregation scheme, not as an order-invariant replacement for the mean score.

\section{Experimental Setup}
\label{sec:setup}

\paragraph{Agents.}
\begin{sloppypar}\noindent We instantiate \method with $K=12$ LLMs accessed through public APIs: \texttt{gpt-5.2}, \texttt{gpt-5-mini}, \texttt{claude-opus-4-5}, \texttt{claude-sonnet-4-5}, \texttt{gemini-3-pro-preview}, \texttt{gemini-3-flash-preview}, \texttt{grok-4-1-fast}, \texttt{deepseek-chat}, \texttt{llama-4-maverick}, \texttt{sonar-pro}, \texttt{kimi-k2.5}, and \texttt{mistral-large}~\cite{openai_models_2025,anthropic_models_2025,google_ai_models_2025,xai_api_2025,deepseek_api_2025,meta_llama_2025,perplexity_sonar_2025,moonshot_kimi_2025,mistral_models_2025}. Provider-native browsing is disabled; \texttt{sonar-pro}, despite being retrieval-native in some deployments, receives the same injected retrieval context as every other agent.\end{sloppypar}

\paragraph{Endogenous task set.}
Each agent generates $n=35$ question instances, giving $N=420$ instances across five categories: factual knowledge, reasoning/logic, current events, creative/open-ended, and practical how-to. The generation prompt does not impose per-category quotas, so category sizes range from 77 to 88 questions and a given generator contributes as few as five questions to some categories. Category-level comparisons are therefore descriptive rather than balanced benchmark estimates.

\paragraph{Labeled benchmarks.}
For an external correctness check, we apply the same shuffle+blind peer-judging procedure to fixed subsets of 264 TruthfulQA validation questions~\cite{lin2021truthfulqa} and 611 GSM8K medium/hard problems~\cite{cobbe2021gsm8k}. TruthfulQA is evaluated with its benchmark accuracy measure; GSM8K uses exact-match numeric accuracy, rescaled to 0--10 only when plotted against the 1--10 peer-score scale. The same item subsets are used for every agent. Because these are labeled, pre-existing benchmarks rather than endogenous tasks, we use them to test alignment with objective correctness, not to establish validity on the open-ended task distribution generated by \method.

\paragraph{Judging and statistical unit.}
Because every agent answers all 420 generated question instances and each answer is scored by 11 non-self peers, the shuffle+blind ranking aggregates $12\times420\times11=55{,}440$ non-self judgments. Primary 95\% confidence intervals are computed by nonparametric bootstrap over question instances, resampling each question together with all attached respondent and judge scores. This preserves within-question dependence, but the corpus audit later identifies repeated and paraphrased tasks, so these intervals may be optimistic relative to a bootstrap clustered by distinct task. Category-level slices are smaller and unbalanced; additional caveats are given in Appendix~\ref{app:category}.

\paragraph{Runtime and scalability.}
In the reported run, question generation took 1 minute 52 seconds, web-grounded answering 34 minutes 45 seconds, and peer judging 9.4 hours: approximately 2.7 hours each for shuffle-only and blind-only and 3.9 hours for shuffle+blind. Dense peer judging requires $O(K^2 n)$ judge calls per regime, but judge--question calls are mutually independent and can be parallelized. Sparse judging or sampled peer panels are natural lower-call variants for larger cohorts.

\paragraph{Artifact availability and appendices.}
All API calls were made in January 2026; results can change as providers update hosted models. The appendices provide the exact prompts and constants (Appendix~\ref{app:prompts}), category-analysis caveats (Appendix~\ref{app:category}), detailed judge-behavior results (Appendix~\ref{app:generosity}), the full benchmark-overlap procedure and sensitivity analysis (Appendix~\ref{app:overlap}), and the item-level human audit (Appendix~\ref{app:audit}). 
Code, raw scores, and audit annotations accompany the submission materials.

\section{Results}
\label{sec:results}

\begin{figure*}[t]
  \centering
  \begin{minipage}[t]{0.48\textwidth}
    \centering
    \includegraphics[width=\linewidth]{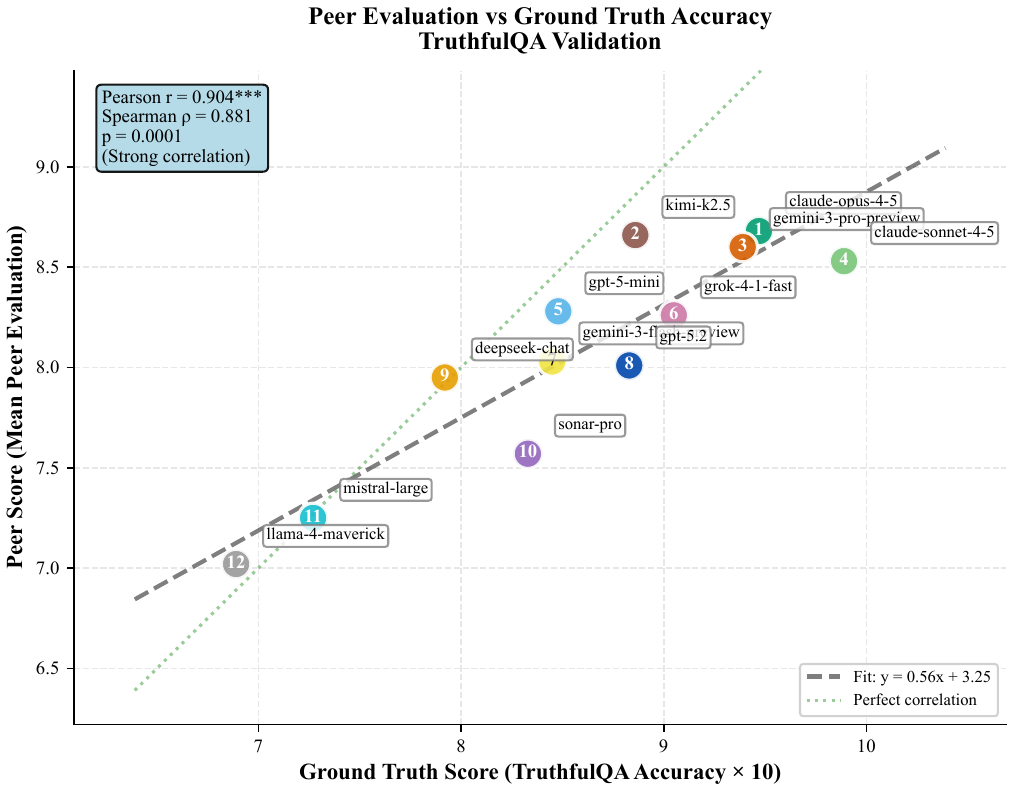}\\[-1pt]
    {\small (a) TruthfulQA: peer score vs.\ accuracy}
  \end{minipage}\hfill
  \begin{minipage}[t]{0.48\textwidth}
    \centering
    \includegraphics[width=\linewidth]{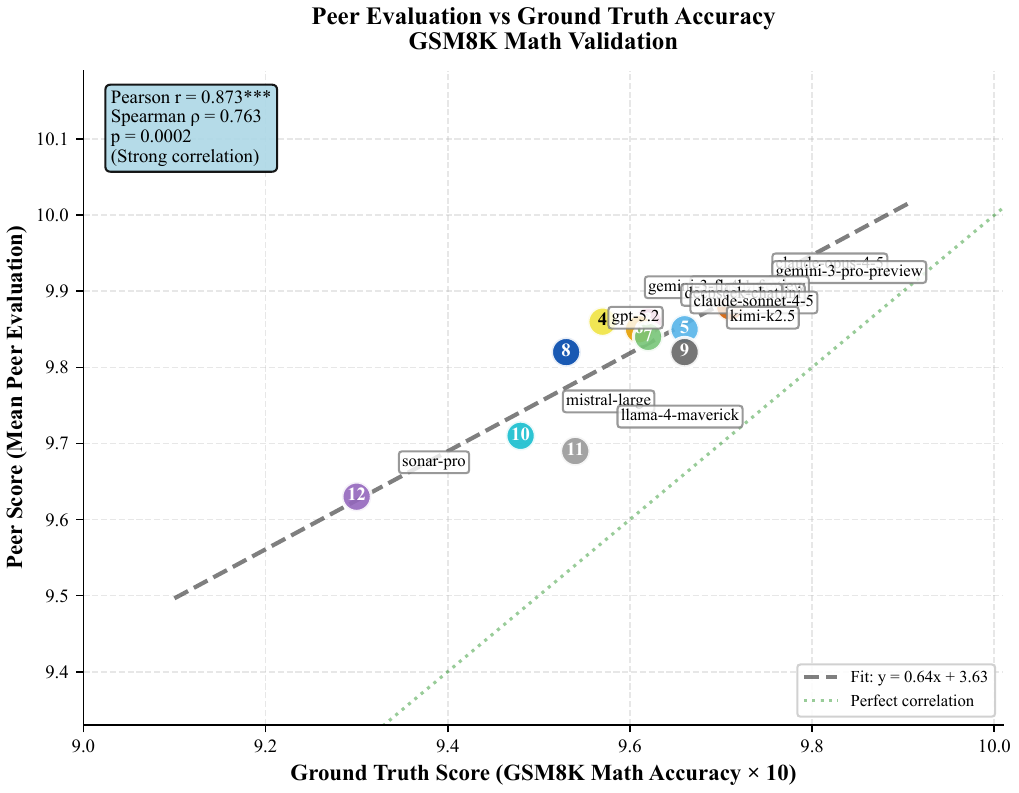}\\[-1pt]
    {\small (b) GSM8K: peer score vs.\ accuracy}
  \end{minipage}
  \caption{Alignment with labeled benchmarks (12 agents, shuffle+blind). (a) TruthfulQA: Pearson $r=\truthfulQAPeerPearson$, Spearman $\rho=0.881$, $p=0.0001$. (b) GSM8K ($10\times$ accuracy): Pearson $r=0.873$, Spearman $\rho=0.763$.}
  \label{fig:validity}
\Description{Two-panel external-validity figure: (a) TruthfulQA peer score versus labeled accuracy and (b) GSM8K peer score versus accuracy.}
\end{figure*}

\subsection{Do Peer Scores Align with Ground Truth?}
\label{sec:validity}
Peer evaluation is relative by construction: a cohort could agree with itself while failing to track correctness. On TruthfulQA, peer scores align strongly with labeled accuracy across the 12 agents (Pearson $r=\truthfulQAPeerPearson$, $p=0.0001$; Spearman $\rho=0.881$, $p<10^{-4}$; Figure~\ref{fig:validity}a). On GSM8K, peer scores also align with exact-match accuracy (Pearson $r=0.873$, $p=0.0002$; Spearman $\rho=0.763$; Figure~\ref{fig:validity}b), extending the check to mathematical reasoning. GSM8K accuracy is concentrated near the upper end of the range, so the corresponding linear correlation should be interpreted under restricted score variance.

\paragraph{Peer versus self evaluation.}
On TruthfulQA, the correlation between peer score and accuracy is substantially higher than the correlation between self-score and accuracy ($r=\truthfulQAPeerPearson$ versus $r=0.538$). This comparison supports excluding self-ratings from the primary ranking: self-evaluation contains useful signal, but in this benchmark it is less aligned with labeled correctness than cross-evaluation by peers.

\paragraph{Scope of this evidence.}
These checks establish alignment with externally defined correctness, but do not validate peer judgments on the endogenous open-ended task set; benchmark familiarity may also inflate alignment. Broader human adjudication with multiple independent annotators would provide a complementary validity check.

\subsection{How Large Are the Measured Bias Effects?}
\label{sec:bias}

We next examine peer assessment on the 420 agent-generated questions; the final subsection audits this task set's overlap, redundancy, and answer quality.

\begin{figure}[t]
    \centering
    \includegraphics[width=\columnwidth]{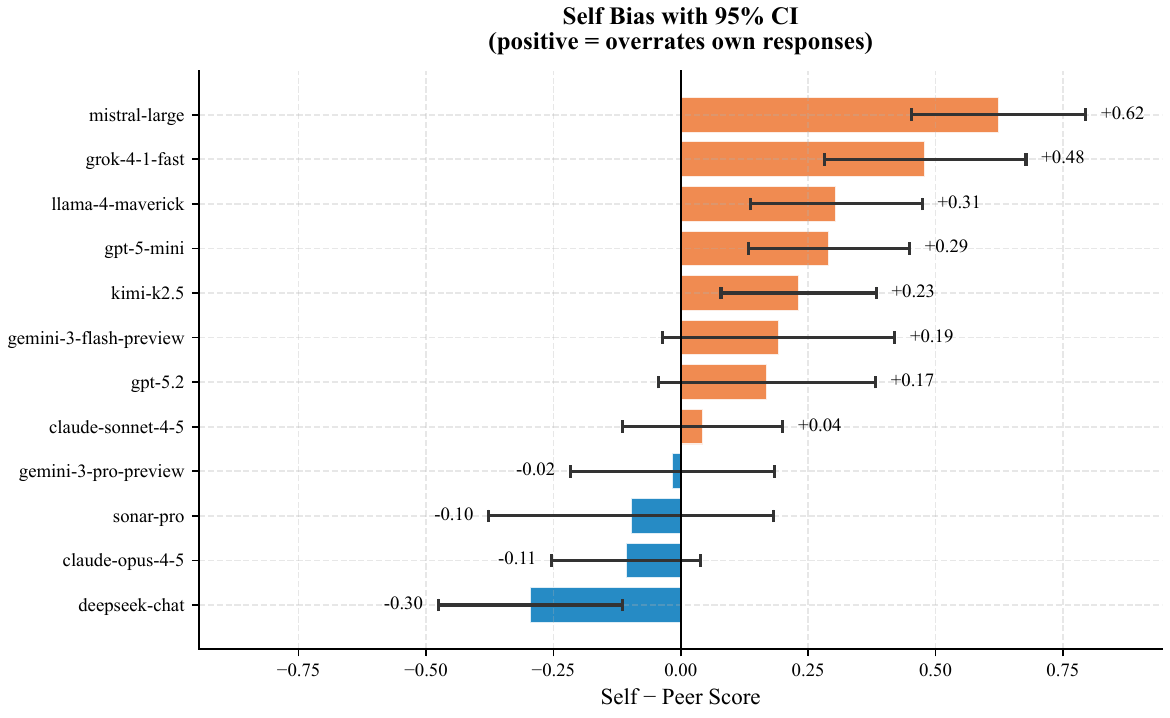}
    \caption{Self bias $\Delta^{\mathrm{self}}_j$ (self score minus peer score) with 95\% bootstrap CIs. Positive values indicate that self-ratings exceed peer ratings of the same answers.}
    \label{fig:self-bias}
\Description{Self-bias estimates for the evaluated agents, defined as self score minus peer score, with 95 percent bootstrap confidence intervals; positive values indicate higher self-ratings.}
\end{figure}

\begin{figure}[t]
    \centering
    \includegraphics[width=\columnwidth]{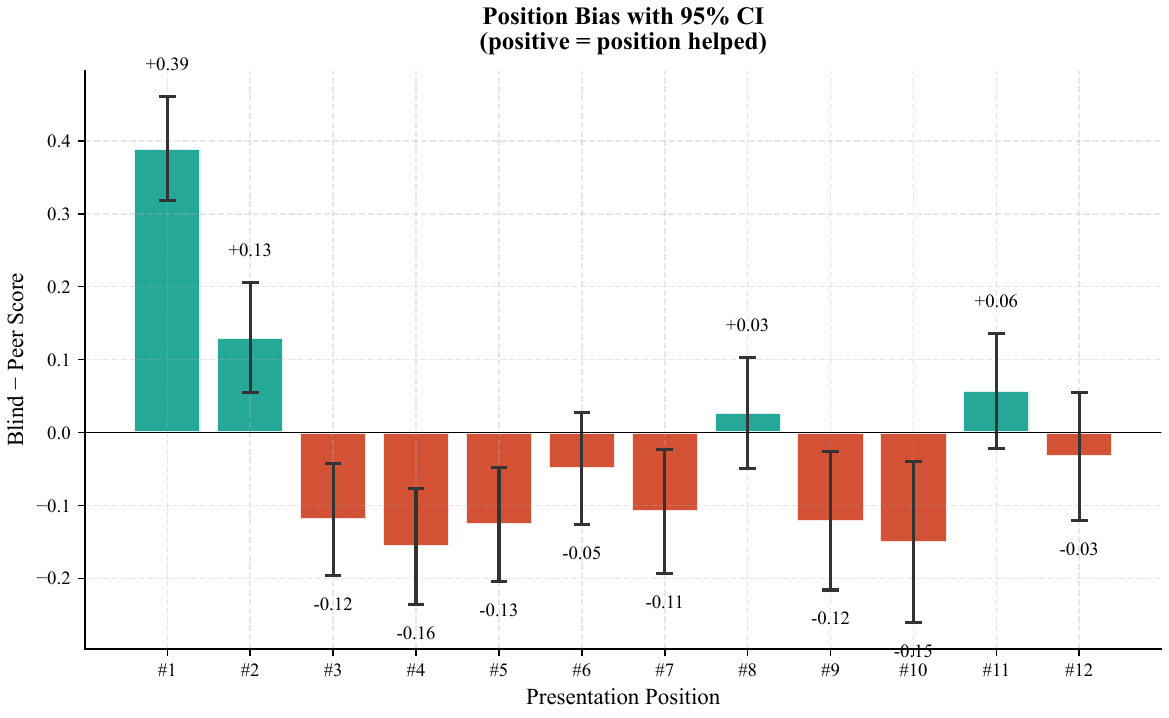}
    \caption{Presentation-position sensitivity. Difference between fixed-position blind judging and the shuffle+blind baseline by answer slot, with 95\% CIs. Positive values indicate higher scores associated with the corresponding presentation position.}
    \label{fig:presentation-position-sensitivity}
\Description{Difference between fixed-position blind judging and the shuffle-and-blind baseline by answer slot, with 95 percent confidence intervals; positive values indicate higher scores associated with the corresponding presentation position.}
\end{figure}

\begin{figure}[t]
    \centering
    \includegraphics[width=\columnwidth]{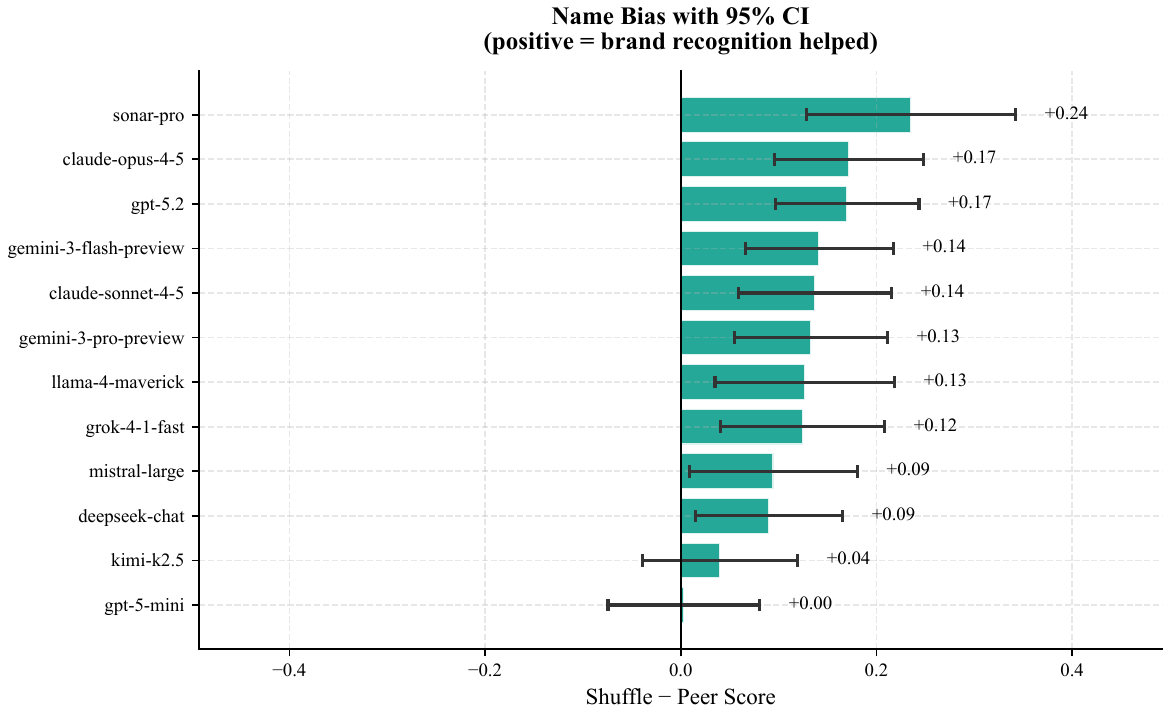}
    \caption{Identity effects: effect of visible model identity, with 95\% CIs. Positive values indicate score inflation when identities are shown.}
    \label{fig:identity-effects}
\Description{Per-agent identity-effect estimates with 95 percent confidence intervals. Positive values indicate higher scores when model identities are visible.}
\end{figure}

\paragraph{Self bias, identity effects, and presentation-position sensitivity.}
Self bias is typically positive: most agents assign their own answers higher scores than their peers do, consistent with documented self-preference~\cite{self_preference_bias_2024}, although several agents are near zero or negative (Figure~\ref{fig:self-bias}). Identity effects also vary by respondent: revealing model names raises some agents' scores and lowers others' (Figure~\ref{fig:identity-effects}). Under fixed answer ordering, the respondent assigned to the first slot received a mean peer score $0.39$ points higher than under shuffle+blind judging. The largest negative difference among respondents assigned to later slots was $-0.16$ points (Figure~\ref{fig:presentation-position-sensitivity}). Because each fixed slot corresponds to one respondent, these contrasts measure respondent-specific sensitivity to the presentation regime and do not identify a general causal effect of answer position. Additional judge-behavior detail is in Appendix~\ref{app:generosity}.

\paragraph{Home-question advantage.}
We compare each agent's peer score on self-authored versus other-authored question instances, excluding self-ratings in both cases. The mean home difference is small ($+0.039$ points; mean Cohen's $d\approx0.02$), but some agents show larger deviations in both directions: \texttt{gemini-3-pro-preview} is $+0.41$ points higher on its own questions ($d=+0.23$), whereas \texttt{grok-4-1-fast} is $-0.47$ points lower ($d=-0.24$). Because judgments are clustered within questions, we treat these values as descriptive and do not attach per-judgment significance claims; per-agent inference requires a question-clustered analysis.

\subsection{How Heterogeneous Are the Judges?}
\label{sec:heterogeneity}

\begin{figure}[t]
    \centering
    \includegraphics[width=\columnwidth]{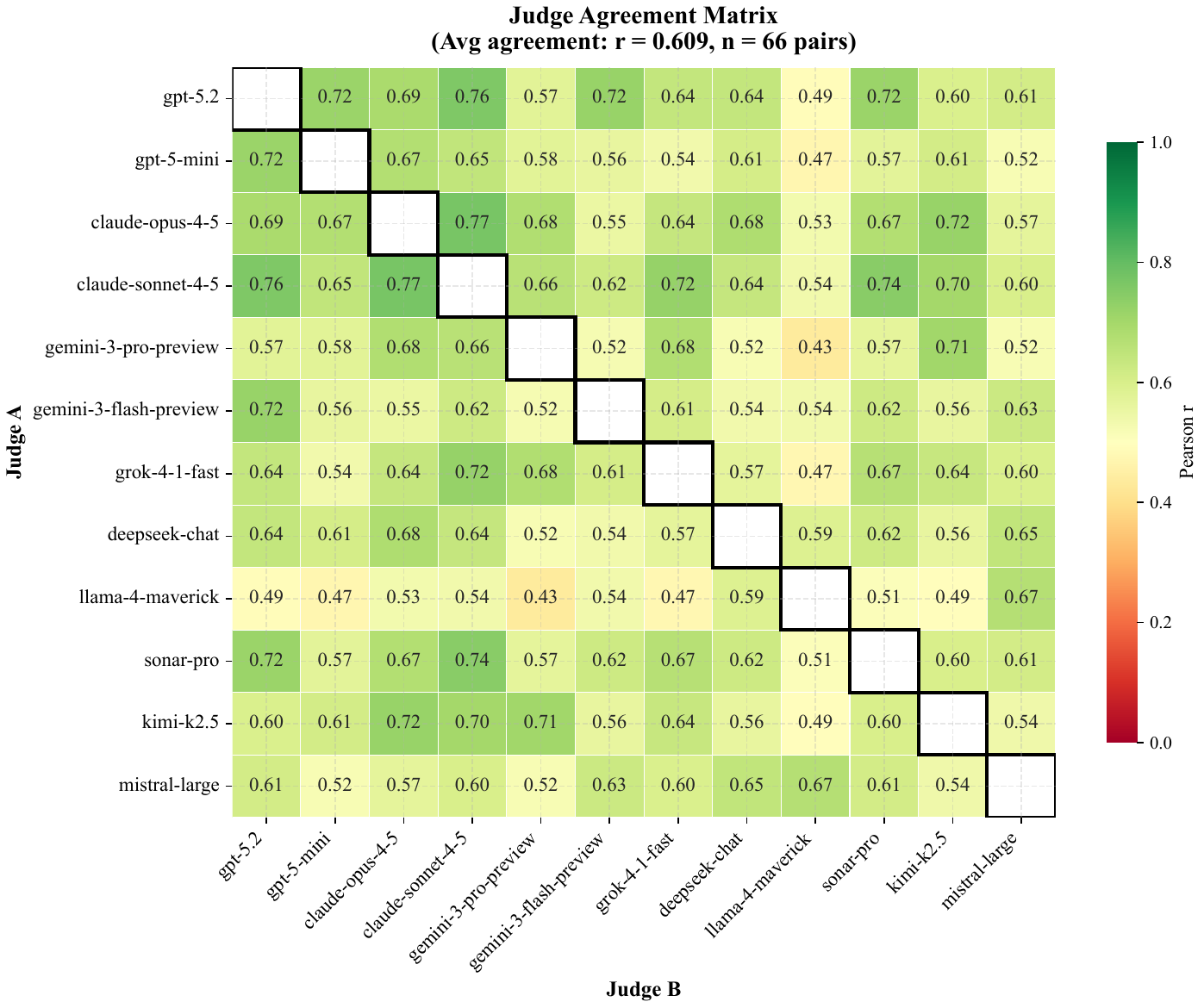}
    \caption{Evaluator consistency across the \method cohort. Each cell reports the Pearson correlation between two agents' score vectors over the same evaluated responses. Pairwise agreement is moderate on average ($\bar{r}=0.609$, 66 judge pairs) but varies substantially across judge pairs, showing that evaluators are not interchangeable even under a shared rubric.}
    \label{fig:judge-agreement}
\Description{Matrix of pairwise Pearson correlations between the 12 agents' score vectors over the same evaluated responses; average agreement is 0.609 across 66 judge pairs, with substantial variation across pairs.}
\end{figure}

\paragraph{Consistency.}
Pairwise correlations between judges' score vectors show moderate average agreement ($\bar{r}=0.609$ over 66 judge pairs), but substantial variation across pairs (Figure~\ref{fig:judge-agreement}). Agreement should therefore be interpreted as a reliability statistic rather than a validity statistic; correlated judges may share preferences, training data, or systematic errors.

\begin{figure}[t]
    \centering
    \includegraphics[width=\columnwidth]{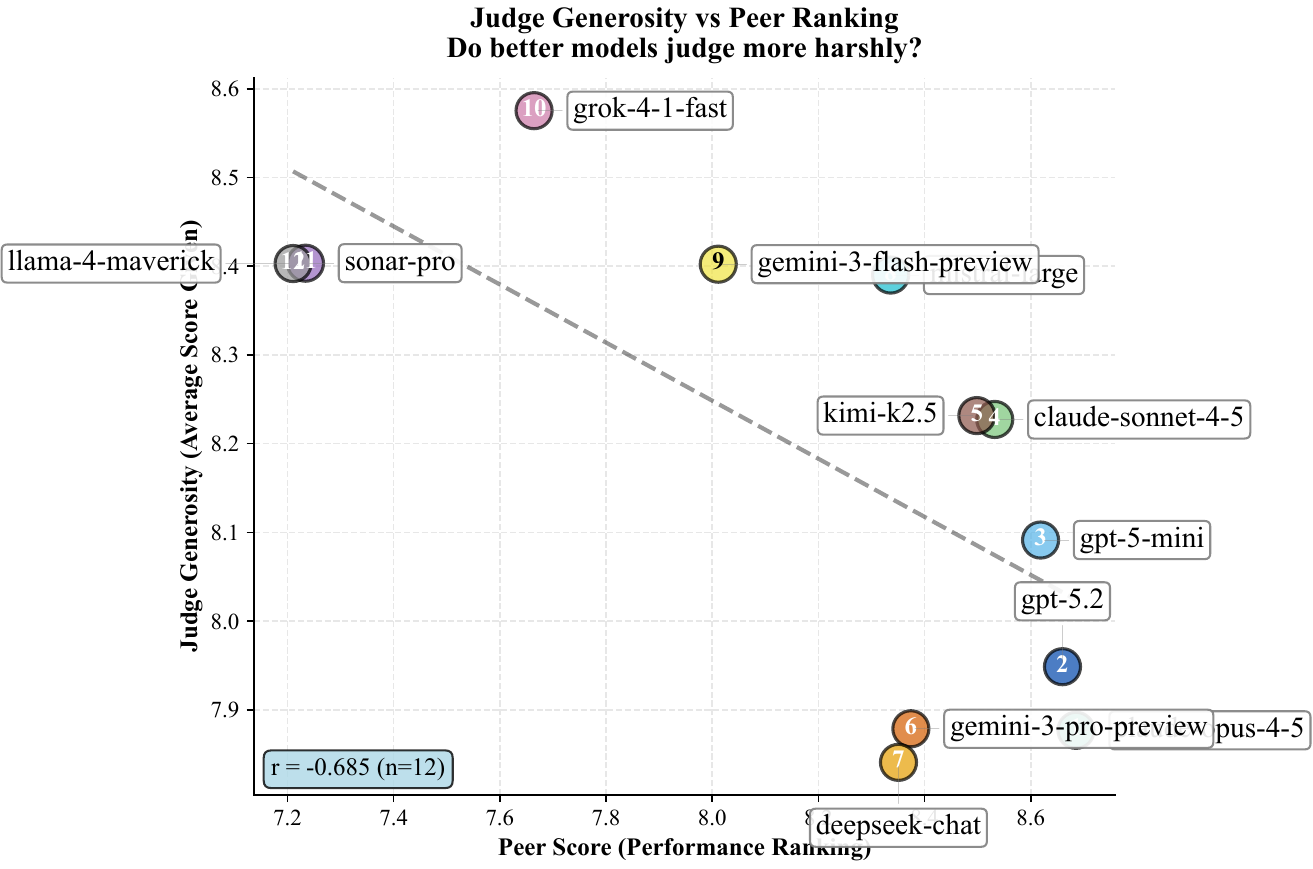}
    \caption{Judge generosity versus peer score across the 12 agents ($r=-0.685$). The association is descriptive; with $n=12$, it should not be read as evidence of a general relationship between model quality and judge harshness.}
    \label{fig:generosity-vs-peer}
\Description{Scatter of judge generosity against peer score for the 12 agents, annotated with r equals -0.685 and n equals 12.}
\end{figure}

\paragraph{Severity.}
Judges also differ in generosity $G_i$. Across the 12 agents, generosity is negatively associated with peer score ($r=-0.685$, Figure~\ref{fig:generosity-vs-peer}), but with $n=12$ we treat this as descriptive rather than evidence that stronger models are intrinsically harsher judges. We therefore test sensitivity to judge-specific score calibration.

\begin{figure}[t]
    \centering
    \includegraphics[width=0.75\columnwidth]{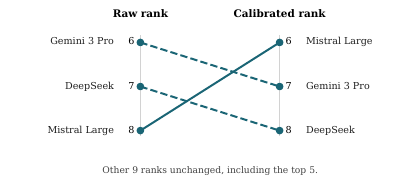}
    \caption{Raw and calibrated ranks; highlighted lines indicate changes ($\rho=0.979$).}
    \label{fig:judge-calibration-ranks}
\Description{Raw and judge-standardized ranks for twelve models under shuffle+blind judging. Mistral Large moves from eighth to sixth, Gemini 3 Pro from sixth to seventh, and DeepSeek Chat from seventh to eighth. All other ranks are unchanged.}
\end{figure}

\paragraph{Judge-calibration sensitivity.}
We assessed shuffle+blind aggregation using judge-wise z-standardization
(Figure~\ref{fig:judge-calibration-ranks}) and an additive model with
answer-specific effects and judge-specific severity offsets, excluding
self-ratings. Both preserved the top-five order and closely matched the
raw ranking (Spearman $\rho=0.979$ and $0.993$, respectively), with maximum
shifts of two and one positions. On the 257 fully scored questions,
z-standardization left the ranking unchanged ($\rho=1.000$). These checks
assess aggregation sensitivity without establishing factual correctness.

\subsection{How Stable Is the Cohort-Relative Ranking?}
\label{sec:rankings}
\begin{figure}[t]
    \centering
    \includegraphics[width=\columnwidth]{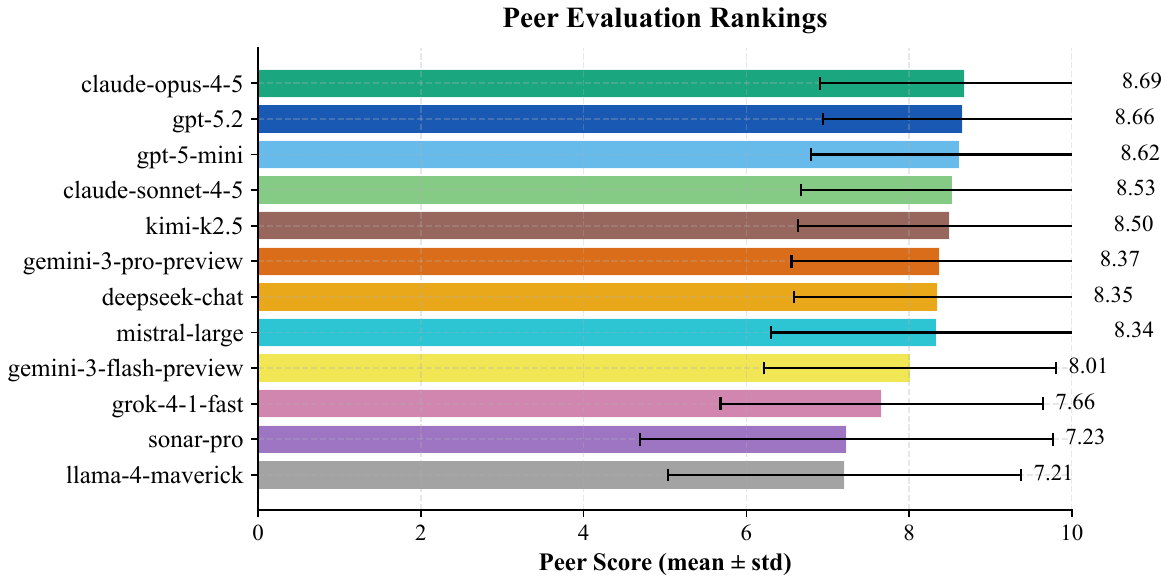}
    \caption{Final \method ranking (shuffle+blind, self-ratings excluded): mean peer score $P_j$ with judgment-dispersion bars ($\pm1$ SD).}
    \label{fig:peer-rankings}
\Description{Final Agents-as-Peers mean peer scores in the shuffle-and-blind regime, excluding self-ratings; error bars show one standard deviation of judgment scores.}
\end{figure}

Figure~\ref{fig:peer-rankings} shows the shuffle+blind ranking with self-ratings excluded. Each bar is the mean peer score $P_j$ over integer 1--10 judgments; error bars show raw judgment dispersion ($\pm1$ SD), not confidence intervals. The two highest means are close relative to this dispersion, while separation is larger among several lower-scoring agents. The figure should therefore be read as a cohort-specific score profile rather than as evidence that every adjacent rank is statistically distinguishable. The category breakdown below is exploratory; additional interpretation caveats are in Appendix~\ref{app:category}.

\begin{figure}[t]
    \centering
    \includegraphics[width=\columnwidth]{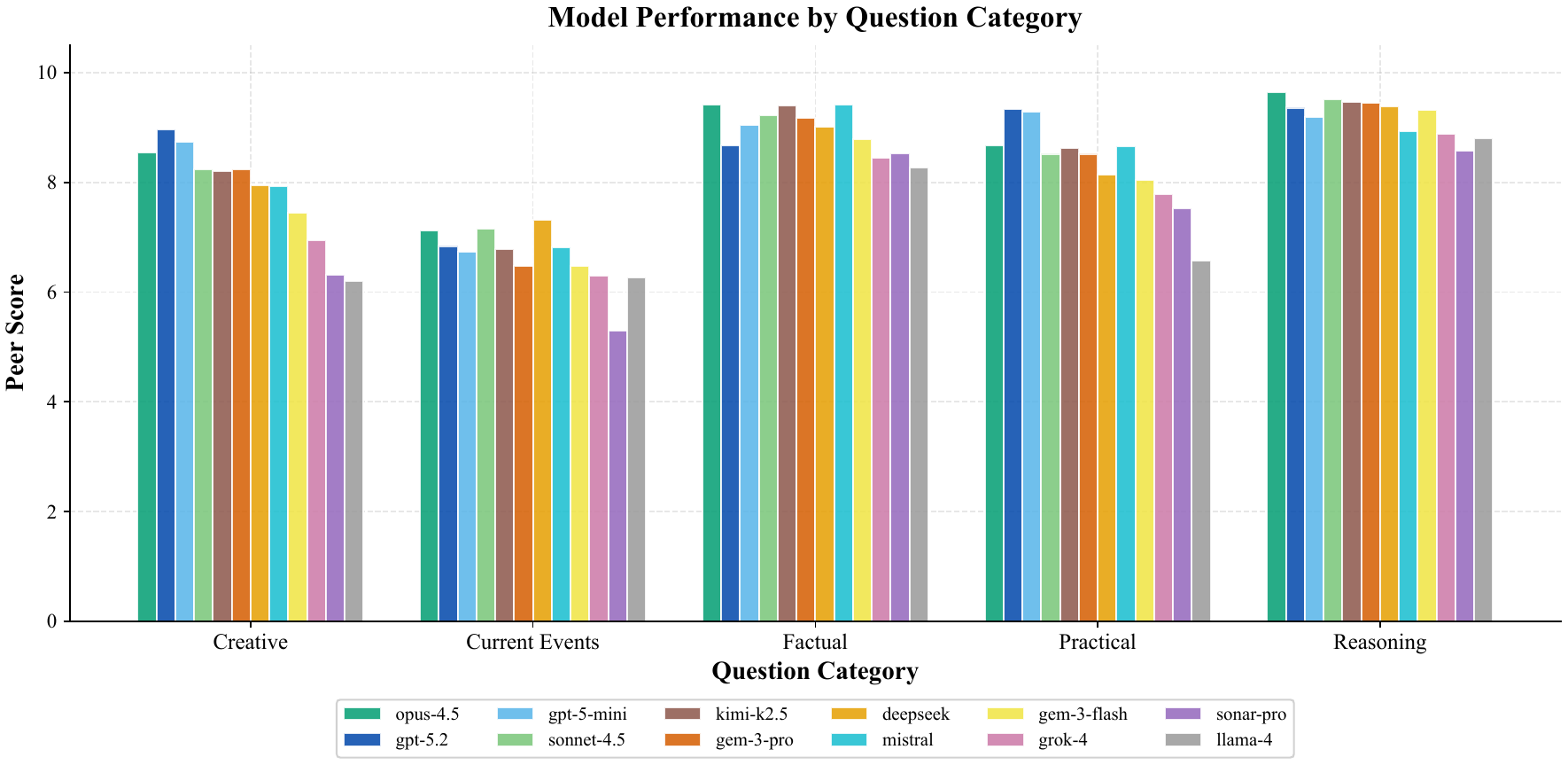}
    \caption{Exploratory breakdown of mean peer scores by question category under the shuffle+blind regime. Category sizes are 77--88 questions and were not balanced per generator, so the figure is descriptive rather than a fully powered set of per-category benchmarks.}
    \label{fig:category-rankings}
\Description{Mean peer scores broken down by question category in the shuffle-and-blind regime; category-level results are exploratory because category sizes are smaller and not balanced per generator.}
\end{figure}

\paragraph{Task-category breakdown (exploratory).}
Aggregate scores can mask specialization by task type. Figure~\ref{fig:category-rankings} breaks down mean peer scores by question category under the same shuffle+blind regime, distinguishing, for example, open-world current-events performance from closed-world factual and reasoning performance. We treat these comparisons as exploratory because the generation prompt did not enforce per-category quotas: categories contain only 77--88 questions, with as few as five questions from a given generator in a category. The current-events category also mixes 35 genuinely retrieval-dependent items with 19 pre-cutoff and 23 open-ended items, so its category-level score should not be interpreted as a pure post-cutoff retrieval benchmark. Additional category-analysis caveats are in Appendix~\ref{app:category}, and the item-level audit underlying the current-events counts is in Appendix~\ref{app:audit}.

\begin{figure}[t]
    \centering
    \includegraphics[width=\columnwidth]{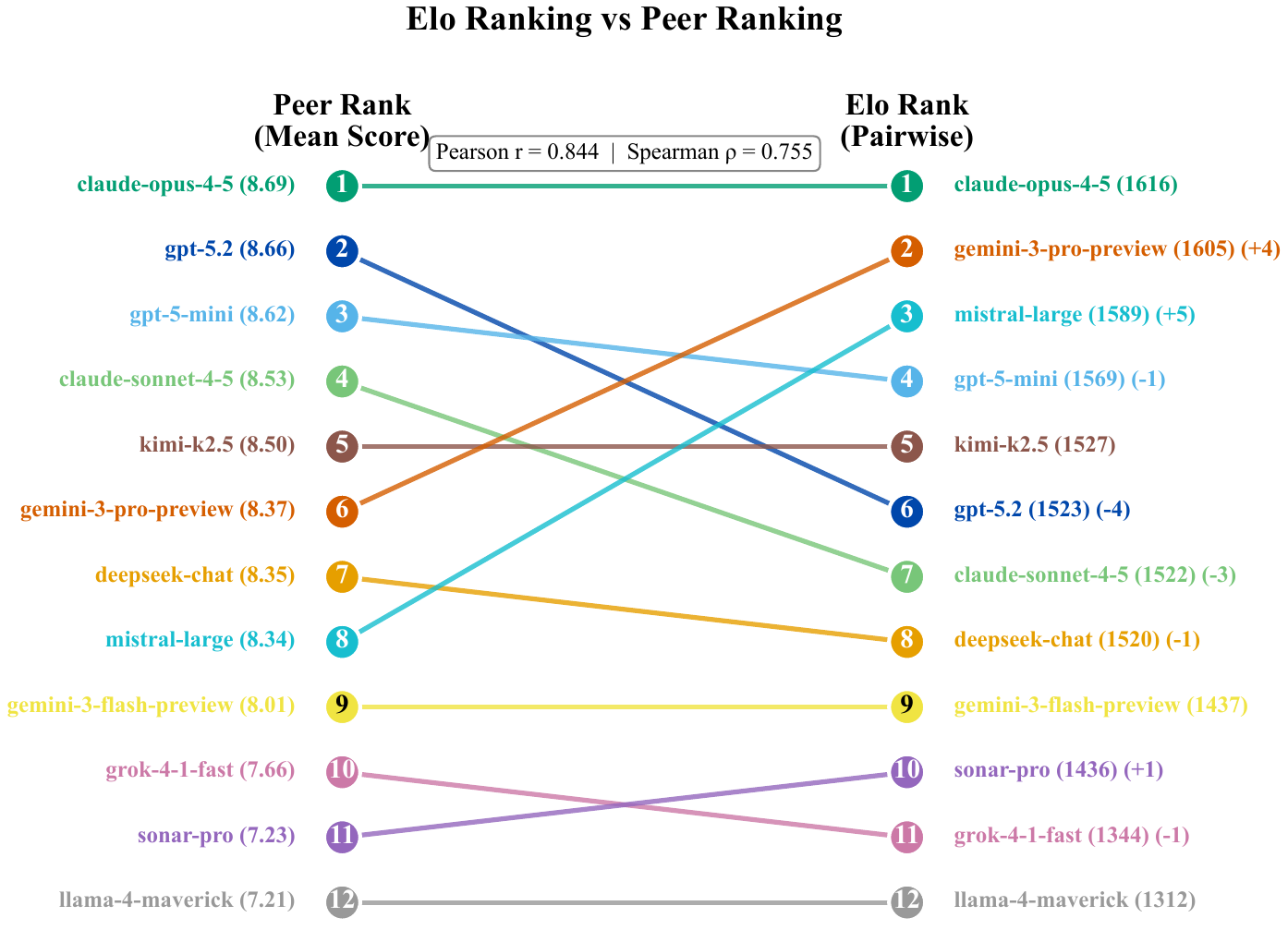}
    \caption{Elo ratings computed from pairwise outcomes show substantial but non-identical agreement with the mean-score ranking (Pearson $r=0.844$; Spearman $\rho=0.755$); some agents shift rank between schemes.}
    \label{fig:elo-vs-peer}
\Description{Comparison of Elo ratings from pairwise outcomes with the mean-score ranking, illustrating agreement and differences between the two ranking schemes.}
\end{figure}

\paragraph{Aggregation sensitivity.}
Elo ratings derived from the same non-self evaluations show substantial but non-identical agreement with mean-score aggregation (Pearson $r=0.844$; Spearman $\rho=0.755$; Figure~\ref{fig:elo-vs-peer}). Because sequential Elo is order-sensitive, this is a robustness check rather than an alternative ground-truth ranking; exact positions remain partly aggregation-dependent.

\subsection{What Does the Generated Task Set Contain?}
\label{sec:corpus}
Because the corpus is used without curation, we characterize it with two complementary audits (full details in Appendix~\ref{app:overlap} and Appendix~\ref{app:audit}).

\paragraph{Benchmark overlap.}
Screening all 420 generated questions against MMLU, TruthfulQA, and GSM8K found limited direct benchmark reproduction. We combined semantic and lexical similarity measures and manually inspected all flagged matches. Across the three benchmark screens, manual inspection identified one verbatim match and four non-verbatim template or topic matches. Table~\ref{tab:overlap-main} summarizes the screening results; Appendix~\ref{app:overlap} provides the detection rules, flagged pairs, sanity checks, and threshold-sensitivity analysis. These findings concern overlap with the three screened benchmarks, not novelty in general, and cannot establish whether the models encountered these benchmarks during training.

\begin{table}[t]
\centering
\caption{Benchmark overlap for 420 generated questions (screening rules in Appendix~\ref{app:overlap}). Duplicate flags are candidates, not verified duplicates. Mean/max summarize maximum embedding cosines; \emph{Below med.} is the share below the benchmark's median within-set nearest-neighbor similarity.}
\label{tab:overlap-main}
\scriptsize
\setlength{\tabcolsep}{2pt}
\begin{tabularx}{\columnwidth}{@{}>{\raggedright\arraybackslash}Xrrrrrrr@{}}
\toprule
\textbf{Benchmark} & \textbf{Items} & \textbf{Mean} & \textbf{Max} & \shortstack{\textbf{Dup.}\\\textbf{flag}} & \textbf{Rev.} & \textbf{Clear} & \shortstack{\textbf{Below}\\\textbf{med.}} \\
\midrule
MMLU       & 14{,}042 & 0.490 & 0.894 & 1 & 9 & 410 & 91.4\% \\
TruthfulQA & 817      & 0.377 & 1.000 & 2 & 0 & 418 & 90.2\% \\
GSM8K      & 1{,}319  & 0.309 & 0.865 & 2 & 0 & 418 & 97.9\% \\
\bottomrule
\end{tabularx}
\end{table}

\paragraph{Human audit.}
A single annotator (one of the authors) assessed question redundancy, current-events temporal status, and a subset of model answers. Across all 420 questions, 11 strings appeared verbatim across two to six generators (33 instances, including 22 redundant copies), and 36 further clusters paraphrased the same task with an identical answer key. Merging exact and paraphrase duplicates leaves 346 distinct tasks. Redundancy is highest in reasoning/logic (38\%) and factual knowledge (23\%) and lowest in creative and current-events questions (2\% and 8\%). All respondents face the same repeated instances, but duplication reweights the task distribution toward repeated problems and reduces effective task diversity; question-instance bootstrap intervals should be interpreted accordingly.

For answer quality, the same annotator assessed 600 responses to 50 questions from 12 models. One question left ``[a specified country or region]'' unspecified, making its 12 responses unassessable. Of the remaining 588 responses, 531 (90.3\%) scored at least 7 out of 10, indicating mostly correct answers with occasional minor errors or omissions. The other 57 (9.7\%) scored below 7: 34 had incorrect or incomplete current-events information, including 11 that accepted a false conference premise; 7 had logical or mathematical errors; 9 had unsupported claims or inconsistencies with creative-task premises; and 7 had procedural, technical, or safety problems.

To compare human and peer assessments, we matched the frozen audit ratings with mean non-self peer scores on 45 questions (540 responses), excluding the unassessable question and four with repeated texts that prevented unambiguous matching. Agreement was moderate (Pearson $r=0.520$; Spearman $\rho=0.577$), providing a check on endogenous responses. This single-annotator assessment of a subset does not establish validity across the task set.

\paragraph{Current events.}
A single annotator (one of the authors) classified the 77 current-events questions: 19 concern events from 2023--2024 and therefore cannot be assumed to require post-training retrieval for models evaluated in January 2026; 23 more are open-ended prompts without a determinate answer. The remaining 35 are determinate and judged to depend on live information. The category as a whole is therefore \emph{not} a pure post-cutoff retrieval benchmark, and we do not interpret its category-level score as evidence specifically about the 35-item retrieval-dependent subset.

\section{Discussion}
\label{sec:discussion}

\paragraph{Peer assessment as a multi-agent evaluation mechanism.}
Our results support peer assessment as a useful but task-dependent evaluation signal. Peer scores track model-level accuracy on two labeled benchmarks, and the cohort ranking changes little after judge calibration. However, agreement with human ratings on generated responses is moderate, and changing the aggregation method alters some ranks. These findings concern different aspects of evaluation: benchmark alignment supports comparisons between models on labeled tasks, while the human audit examines judgments of individual generated responses. Neither establishes correctness across the full generated corpus. From a multi-agent perspective, the same agents jointly define the task distribution, compete as respondents, and assess their peers: the collective ranking emerges from these agent-to-agent judgments. Role symmetry distributes evaluative authority, but leaves the outcome dependent on the cohort's shared preferences and errors. The protocol makes these dependencies available for systematic analysis, so rankings should be interpreted together with the task audit and evaluator diagnostics.

\paragraph{Connection to impartial peer selection.}
Our agents are prompted evaluators rather than strategic optimizers, but the protocol still has the structure of a peer-selection problem. Excluding self-ratings prevents an agent from directly contributing to its own peer score, but does not guarantee impartial selection: its ratings of competitors can still affect its relative rank. Classic impartial-selection mechanisms are designed for stronger strategic settings in which agents may optimize their reports~\cite{holzman2013impartial,alon2011sum,kurokawa2015impartial,aziz2019strategyproof}. If LLMs are trained or prompted to optimize against peer-evaluation leaderboards, partition-based impartial mechanisms or peer-prediction ideas~\cite{miller2005eliciting} become natural next steps. The present bias measurements provide an empirical baseline for testing such mechanisms.

\paragraph{Scalability.}
Dense judging produces $NK(K-1)=K^2(K-1)n$ non-self
scores per regime. If each judge scores all $K$ answers
to a question in one call, the number of judge calls
is $KN=K^2n$. The calls are parallelizable, but larger cohorts will likely require sampled judge panels, sparse comparison graphs, or adaptive allocation of judging effort. Those variants introduce a new statistical question---how much judging coverage is needed to preserve ranking and bias estimates---that the current dense design does not address.

\section{Limitations and Ethical Considerations}
\label{sec:limitations}

\paragraph{Cohort dependence.}
\method measures relative performance within a cohort, not absolute capability. Because the cohort determines both questions and judges, separate runs require shared anchors, items, or calibration for comparison.

\paragraph{Shared blind spots and external validity.}
A cohort may agree while sharing errors, training data, or preferences, so inter-judge agreement indicates reliability, not correctness. TruthfulQA and GSM8K establish alignment with labeled accuracy only on those benchmarks, not validity on endogenous open-ended tasks. Repeated and paraphrased tasks also reduce effective diversity and may make question-level bootstrap intervals optimistic.

\paragraph{Interpretation of bias contrasts.}
The bias contrasts depend on our interventions. Model style may weaken identity blinding, and fixed answer slots confound position with respondent characteristics; we therefore report presentation-position sensitivity, not a causal position effect. Excluding self-ratings also does not prevent agents from affecting their rank by judging competitors.

\paragraph{Task-generation scope.}
Endogenous generation does not guarantee broad or novel coverage. Agents converge on canonical problems, and 42 of 77 current-events questions are pre-2025 or open-ended. Untested capabilities remain outside the measurement, and the overlap screen cannot rule out training contamination.

\paragraph{Ethical considerations.}
We evaluate models through public APIs under their usage policies; no personal data or external human-subject data are used. TruthfulQA (Apache 2.0) and GSM8K (MIT) are used solely for evaluation. Because model rankings can influence deployment decisions, \method should not be used as the sole basis for high-stakes model selection: the outputs are time-stamped, cohort-dependent, and partly defined by model-generated judgments. Web retrieval can surface inaccurate material, and fluent but incorrect answers can still receive high scores. Generative assistants (Claude Code, ChatGPT, and Claude) were used for experiment orchestration, script editing, and manuscript editing; the authors specified and verified the study design, analyses, results, and interpretations.

\section{Conclusion}

\method casts LLM evaluation as a multi-agent peer assessment problem in which the same cohort generates tasks, answers them, and supplies distributed agent-to-agent judgments under controlled identity and presentation conditions. In our 12-agent study, peer scores align strongly with labeled accuracy on TruthfulQA and GSM8K, and on TruthfulQA they correlate with accuracy more strongly than self-scores do. At the same time, self-evaluation, visible identity, presentation position, and judge severity all measurably affect the scoring process. The generated task set shows little direct overlap with the three screened public benchmarks but substantial internal redundancy, demonstrating that the task distribution itself must be audited alongside the judges. \method does not replace human adjudication or correctness-based benchmarks; it provides a cohort-relative testbed for studying peer assessment, impartiality, and evaluator bias among AI agents.

\bibliographystyle{ACM-Reference-Format}
\bibliography{peerrank3}

@book{elo1978rating,
  title     = {The Rating of Chessplayers, Past and Present},
  author    = {Elo, Arpad E.},
  publisher = {Arco Publishing},
  year      = {1978}
}

@misc{liang2022helm,
  title         = {{HELM}: Holistic Evaluation of Language Models},
  author        = {Liang, Percy and Bommasani, Rishi and Lee, Tony and Tsipras, Dimitris and Soylu, Dilara and Yasunaga, Michihiro and Zhang, Yian and Narayanan, Deepak and Wu, Yuhuai and Kumar, Ananya and others},
  year          = {2022},
  eprint        = {2211.09110},
  archivePrefix = {arXiv},
  primaryClass  = {cs.CL},
  url           = {https://arxiv.org/abs/2211.09110}
}

@article{wang2024mmlu,
  title   = {{MMLU}-Pro: A More Robust and Challenging Multi-Task Language Understanding Benchmark},
  author  = {Wang, Yubo and others},
  journal = {Advances in Neural Information Processing Systems},
  volume  = {37},
  pages   = {95266--95290},
  year    = {2024}
}

@misc{lin2021truthfulqa,
  title         = {{TruthfulQA}: Measuring How Models Mimic Human Falsehoods},
  author        = {Lin, Stephanie and Hilton, Jacob and Evans, Owain},
  year          = {2021},
  eprint        = {2109.07958},
  archivePrefix = {arXiv},
  primaryClass  = {cs.CL},
  url           = {https://arxiv.org/abs/2109.07958}
}

@inproceedings{liu2023agentbench,
  author    = {Liu, Xiao and Yu, Hao and Zhang, Hanchen and Xu, Yifan and Lei, Xuanyu and Lai, Hanyu and Gu, Yu and Ding, Hangliang and Men, Kaiwen and Yang, Kejuan and Zhang, Shudan and Deng, Xiang and Zeng, Aohan and Du, Zhengxiao and Zhang, Chenhui and Shen, Sheng and Zhang, Tianjun and Su, Yu and Sun, Huan and Huang, Minlie and Dong, Yuxiao and Tang, Jie},
  booktitle = {International Conference on Learning Representations},
  editor    = {B. Kim and Y. Yue and S. Chaudhuri and K. Fragkiadaki and M. Khan and Y. Sun},
  pages     = {52989--53046},
  title     = {{AgentBench}: Evaluating {LLM}s as Agents},
  url       = {https://proceedings.iclr.cc/paper_files/paper/2024/file/e9df36b21ff4ee211a8b71ee8b7e9f57-Paper-Conference.pdf},
  volume    = {2024},
  year      = {2024}
}

@inproceedings{zhou2023webarena,
  author    = {Zhou, Shuyan and Xu, Frank F and Zhu, Hao and Zhou, Xuhui and Lo, Robert and Sridhar, Abishek and Cheng, Xianyi and Ou, Tianyue and Bisk, Yonatan and Fried, Daniel and Alon, Uri and Neubig, Graham},
  booktitle = {International Conference on Learning Representations},
  editor    = {B. Kim and Y. Yue and S. Chaudhuri and K. Fragkiadaki and M. Khan and Y. Sun},
  pages     = {15585--15606},
  title     = {{WebArena}: A Realistic Web Environment for Building Autonomous Agents},
  url       = {https://proceedings.iclr.cc/paper_files/paper/2024/file/4410c0711e9154a7a2d26f9b3816d1ef-Paper-Conference.pdf},
  volume    = {2024},
  year      = {2024}
}

@article{self_preference_bias_2024,
  title         = {Self-Preference Bias in {LLM}-as-a-Judge},
  author        = {Wataoka, Koki and Takahashi, Tsubasa and Ri, Ryokan},
  journal       = {arXiv preprint arXiv:2410.21819},
  year          = {2024},
  eprint        = {2410.21819},
  archivePrefix = {arXiv},
  primaryClass  = {cs.CL},
  url           = {https://arxiv.org/abs/2410.21819}
}

@misc{zheng2023judging_llm_as_judge,
  title         = {Judging {LLM}-as-a-Judge with {MT}-Bench and Chatbot Arena},
  author        = {Zheng, Lianmin and Chiang, Wei-Lin and Sheng, Ying and Zhuang, Siyuan and Wu, Zhanghao and Zhuang, Yonghao and Lin, Zi and Li, Zhuohan and Li, Dacheng and Xing, Eric P. and Zhang, Hao and Gonzalez, Joseph E. and Stoica, Ion},
  year          = {2023},
  eprint        = {2306.05685},
  archivePrefix = {arXiv},
  primaryClass  = {cs.CL},
  url           = {https://arxiv.org/abs/2306.05685}
}

@misc{chiang2024chatbot_arena,
  title         = {Chatbot Arena: An Open Platform for Evaluating {LLM}s by Human Preference},
  author        = {Chiang, Wei-Lin and Zheng, Lianmin and Sheng, Ying and Angelopoulos, Anastasios and Li, Zhuohan and Li, Dacheng and Zhu, Banghua and Zhang, Hao and Gonzalez, Joseph E. and Stoica, Ion},
  year          = {2024},
  eprint        = {2403.04132},
  archivePrefix = {arXiv},
  primaryClass  = {cs.CL},
  url           = {https://arxiv.org/abs/2403.04132}
}

@inproceedings{thakur2025judging,
  title={Judging the judges: Evaluating alignment and vulnerabilities in llms-as-judges},
  author={Thakur, Aman Singh and Choudhary, Kartik and Ramayapally, Venkat Srinik and Vaidyanathan, Sankaran and Hupkes, Dieuwke},
  booktitle={Proceedings of the Fourth Workshop on Generation, Evaluation and Metrics (GEM$^2$)},
  pages={404--430},
  year={2025}
}

@inproceedings{zhao2024justice,
  author    = {Ye, Jiayi and Wang, Yanbo and Huang, Yue and Chen, Dongping and Zhang, Qihui and Moniz, Nuno and Gao, Tian and Geyer, Werner and Huang, Chao and Chen, Pin-Yu and Chawla, Nitesh and Zhang, Xiangliang},
  booktitle = {International Conference on Learning Representations},
  editor    = {Y. Yue and A. Garg and N. Peng and F. Sha and R. Yu},
  pages     = {102351--102390},
  title     = {Justice or Prejudice? Quantifying Biases in {LLM}-as-a-Judge},
  url       = {https://proceedings.iclr.cc/paper_files/paper/2025/file/fdca08d371e4b6c031397909e20043bd-Paper-Conference.pdf},
  volume    = {2025},
  year      = {2025}
}

@inproceedings{zhang2025upme,
  title={Upme: An unsupervised peer review framework for multimodal large language model evaluation},
  author={Zhang, Qihui and Ning, Munan and Liu, Zheyuan and Huang, Yue and Yang, Shuo and Wang, Yanbo and Ye, Jiayi and Chen, Xiao and Song, Yibing and Yuan, Li},
  booktitle={Proceedings of the Computer Vision and Pattern Recognition Conference},
  pages={9165--9174},
  year={2025}
}

@inproceedings{shi2025judging,
  title={Judging the judges: A systematic study of position bias in llm-as-a-judge},
  author={Shi, Lin and Ma, Chiyu and Liang, Wenhua and Diao, Xingjian and Ma, Weicheng and Vosoughi, Soroush},
  booktitle={Proceedings of the 14th International Joint Conference on Natural Language Processing and the 4th Conference of the Asia-Pacific Chapter of the Association for Computational Linguistics},
  pages={292--314},
  year={2025}
}

@inproceedings{lambert2025rewardbench,
    title     = {{RewardBench}: Evaluating Reward Models for Language Modeling},
    author    = {Lambert, Nathan and Pyatkin, Valentina and Morrison, Jacob and Miranda, LJ and Lin, Bill Yuchen and Chandu, Khyathi and Dziri, Nouha and Kumar, Sachin and Zick, Tom and Choi, Yejin and Smith, Noah A. and Hajishirzi, Hannaneh},
    booktitle = {Findings of the Association for Computational Linguistics: NAACL 2025},
    month     = apr,
    year      = {2025},
    publisher = {Association for Computational Linguistics},
    address   = {Albuquerque, New Mexico},
    pages     = {1755--1797},
    url       = {https://aclanthology.org/2025.findings-naacl.96},
    doi       = {10.18653/v1/2025.findings-naacl.96}
}

@inproceedings{liu2023geval,
  author    = {Yang Liu and Dan Iter and Yichong Xu and Shuohang Wang and Ruochen Xu and Chenguang Zhu},
  title     = {{G-Eval}: {NLG} Evaluation using {GPT-4} with Better Human Alignment},
  booktitle = {Proceedings of the 2023 Conference on Empirical Methods in Natural Language Processing (EMNLP)},
  editor    = {Houda Bouamor and Juan Pino and Kalika Bali},
  pages     = {2511--2522},
  publisher = {Association for Computational Linguistics},
  year      = {2023},
  doi       = {10.18653/V1/2023.EMNLP-MAIN.153},
  url       = {https://doi.org/10.18653/v1/2023.emnlp-main.153}
}

@article{chu2024pre,
  title={Pre: A peer review based large language model evaluator},
  author={Chu, Zhumin and Ai, Qingyao and Tu, Yiteng and Li, Haitao and Liu, Yiqun},
  journal={arXiv preprint arXiv:2401.15641},
  year={2024}
}

@inproceedings{inger2025forget,
  title={Forget What You Know about LLMs Evaluations-LLMs are Like a Chameleon},
  author={Inger, Nurit Cohen and Elisha, Yehonatan and Shapira, Bracha and Rokach, Lior and Cohen, Seffi},
  booktitle={Proceedings of the 2025 Conference on Empirical Methods in Natural Language Processing},
  pages={21675--21688},
  year={2025}
}

@inproceedings{chen2025benchmarking,
  title={Benchmarking large language models under data contamination: A survey from static to dynamic evaluation},
  author={Chen, Simin and Chen, Yiming and Li, Zexin and Jiang, Yifan and Wan, Zhongwei and He, Yixin and Ran, Dezhi and Gu, Tianle and Li, Haizhou and Xie, Tao and others},
  booktitle={Proceedings of the 2025 Conference on Empirical Methods in Natural Language Processing},
  pages={10091--10109},
  year={2025}
}

@inproceedings{balloccu-etal-2024-leak,
    title = "Leak, Cheat, Repeat: Data Contamination and Evaluation Malpractices in Closed-Source {LLM}s",
    author = "Balloccu, Simone  and
      Schmidtov{\'a}, Patr{\'i}cia  and
      Lango, Mateusz  and
      Dusek, Ondrej",
    editor = "Graham, Yvette  and
      Purver, Matthew",
    booktitle = "Proceedings of the 18th Conference of the European Chapter of the Association for Computational Linguistics (Volume 1: Long Papers)",
    month = mar,
    year = "2024",
    address = "St. Julian{'}s, Malta",
    publisher = "Association for Computational Linguistics",
    url = "https://aclanthology.org/2024.eacl-long.5/",
    doi = "10.18653/v1/2024.eacl-long.5",
    pages = "67--93"
}

@article{rontogiannis2025interactive,
  title = {Interactive Evaluation of Large Language Models for Multi-Requirement Software Engineering Tasks},
  author = {Rontogiannis, Dimitrios and Peyrard, Maxime and Baldwin, Nicolas and Josifoski, Martin and West, Robert and Gunopulos, Dimitrios},
  journal = {arXiv preprint arXiv:2508.18905},
  year = {2025},
  url = {https://arxiv.org/abs/2508.18905}
}

@inproceedings{you2024llmevolve,
  title = {{LLM}-Evolve: Evaluation for {LLM}'s Evolving Capability on Benchmarks},
  author = {You, Jiaxuan and Liu, Mingjie and Prabhumoye, Shrimai and Patwary, Mostofa and Shoeybi, Mohammad and Catanzaro, Bryan},
  booktitle = {Proceedings of the 2024 Conference on Empirical Methods in Natural Language Processing},
  pages = {16937--16942},
  year = {2024},
  publisher = {Association for Computational Linguistics},
  doi = {10.18653/v1/2024.emnlp-main.940}
}

@inproceedings{hendrycks2021measuring,
  title = {Measuring Massive Multitask Language Understanding},
  author = {Hendrycks, Dan and Burns, Collin and Basart, Steven and Zou, Andy and Mazeika, Mantas and Song, Dawn and Steinhardt, Jacob},
  booktitle = {Proceedings of the International Conference on Learning Representations (ICLR)},
  year = {2021},
  url = {https://openreview.net/forum?id=d7KBjmI3Dk}
}

@inproceedings{rajpurkar2016squad,
  title = {{SQuAD}: 100,000+ Questions for Machine Comprehension of Text},
  author = {Rajpurkar, Pranav and Zhang, Jian and Lopyrev, Konstantin and Liang, Percy},
  booktitle = {Proceedings of the 2016 Conference on Empirical Methods in Natural Language Processing},
  pages = {2383--2392},
  year = {2016},
  publisher = {Association for Computational Linguistics},
  doi = {10.18653/v1/D16-1264}
}

@article{cobbe2021gsm8k,
  title = {Training Verifiers to Solve Math Word Problems},
  author = {Cobbe, Karl and Kosaraju, Vineet and Bavarian, Mohammad and Chen, Mark and Jun, Heewoo and Kaiser, Lukasz and Plappert, Matthias and Tworek, Jerry and Hilton, Jacob and Nakano, Reiichiro and Hesse, Christopher and Schulman, John},
  journal = {arXiv preprint arXiv:2110.14168},
  year = {2021},
  url = {https://arxiv.org/abs/2110.14168}
}

@misc{openai_models_2025,
  author       = {{OpenAI}},
  title        = {Models documentation},
  howpublished = {\url{https://platform.openai.com/docs/models}},
  year         = {2025},
  note         = {Accessed: 2026-01-23}
}

@misc{anthropic_models_2025,
  author       = {{Anthropic}},
  title        = {Models overview},
  howpublished = {\url{https://docs.anthropic.com/en/docs/about-claude/models}},
  year         = {2025},
  note         = {Accessed: 2026-01-23}
}

@misc{google_ai_models_2025,
  author       = {{Google}},
  title        = {Gemini models documentation},
  howpublished = {\url{https://ai.google.dev/gemini-api/docs/models}},
  year         = {2025},
  note         = {Accessed: 2026-01-23}
}

@misc{xai_api_2025,
  author       = {{xAI}},
  title        = {xAI API documentation},
  howpublished = {\url{https://docs.x.ai/}},
  year         = {2025},
  note         = {Accessed: 2026-01-23}
}

@misc{deepseek_api_2025,
  author       = {{DeepSeek}},
  title        = {DeepSeek API documentation},
  howpublished = {\url{https://platform.deepseek.com/docs}},
  year         = {2025},
  note         = {Accessed: 2026-01-23}
}

@misc{meta_llama_2025,
  author       = {{Meta}},
  title        = {Llama model documentation},
  howpublished = {\url{https://www.llama.com/docs/}},
  year         = {2025},
  note         = {Accessed: 2026-01-23}
}

@misc{perplexity_sonar_2025,
  author       = {{Perplexity AI}},
  title        = {Sonar API documentation},
  howpublished = {\url{https://docs.perplexity.ai/}},
  year         = {2025},
  note         = {Accessed: 2026-01-23}
}

@misc{moonshot_kimi_2025,
  author       = {{Moonshot AI}},
  title        = {Kimi API documentation},
  howpublished = {\url{https://platform.moonshot.cn/docs/}},
  year         = {2025},
  note         = {Accessed: 2026-01-23}
}

@misc{mistral_models_2025,
  author       = {{Mistral AI}},
  title        = {Mistral models documentation},
  howpublished = {\url{https://docs.mistral.ai/getting-started/models/}},
  year         = {2025},
  note         = {Accessed: 2026-01-23}
}

@misc{loi2025autobench,
      title={AutoBench: Automating LLM Evaluation through Reciprocal Peer Assessment}, 
      author={Dario Loi and Elena Maria Muià and Federico Siciliano and Giovanni Trappolini and Vincenzo Crisà and Peter Kruger and Fabrizio Silvestri},
      year={2025},
      eprint={2510.22593},
      archivePrefix={arXiv},
      primaryClass={cs.CL},
      url={https://arxiv.org/abs/2510.22593}, 
}

@article{piech2013tuned,
  title   = {Tuned Models of Peer Assessment in {MOOCs}},
  author  = {Piech, Chris and Huang, Jonathan and Chen, Zhenghao and Do, Chuong and Ng, Andrew and Koller, Daphne},
  journal = {Proceedings of the 6th International Conference on Educational Data Mining (EDM)},
  year    = {2013},
  note    = {arXiv:1307.2579}
}

@book{lord1980applications,
  title     = {Applications of Item Response Theory to Practical Testing Problems},
  author    = {Lord, Frederic M.},
  year      = {1980},
  publisher = {Lawrence Erlbaum Associates},
  address   = {Hillsdale, NJ}
}

@inproceedings{wright2015mechanical,
  author    = {James R. Wright and Chris Thornton and Kevin Leyton-Brown},
  title     = {Mechanical {TA}: Partially Automated High-Stakes Peer Grading},
  booktitle = {Proceedings of the 46th ACM Technical Symposium on Computer Science Education},
  year      = {2015},
  pages     = {96--101},
  doi       = {10.1145/2676723.2677278}
}

@article{holzman2013impartial,
  author  = {Ron Holzman and Herv{\'e} Moulin},
  title   = {Impartial Nominations for a Prize},
  journal = {Econometrica},
  volume  = {81},
  number  = {1},
  pages   = {173--196},
  year    = {2013},
  doi     = {10.3982/ECTA10523}
}

@inproceedings{alon2011sum,
  author    = {Noga Alon and Felix Fischer and Ariel Procaccia and Moshe Tennenholtz},
  title     = {Sum of Us: Strategyproof Selection from the Selectors},
  booktitle = {Proceedings of the 13th Conference on Theoretical Aspects of Rationality and Knowledge},
  year      = {2011},
  pages     = {101--110},
  doi       = {10.1145/2000378.2000390}
}

@inproceedings{kurokawa2015impartial,
  author    = {David Kurokawa and Omer Lev and Jamie Morgenstern and Ariel D. Procaccia},
  title     = {Impartial Peer Review},
  booktitle = {Proceedings of the 24th International Joint Conference on Artificial Intelligence},
  year      = {2015},
  pages     = {582--588}
}

@article{aziz2019strategyproof,
  author  = {Haris Aziz and Omer Lev and Nicholas Mattei and Jeffrey S. Rosenschein and Toby Walsh},
  title   = {Strategyproof Peer Selection using Randomization, Partitioning, and Apportionment},
  journal = {Artificial Intelligence},
  volume  = {275},
  pages   = {295--309},
  year    = {2019},
  doi     = {10.1016/j.artint.2019.06.004}
}

@article{miller2005eliciting,
  author  = {Nolan Miller and Paul Resnick and Richard Zeckhauser},
  title   = {Eliciting Informative Feedback: The Peer-Prediction Method},
  journal = {Management Science},
  volume  = {51},
  number  = {9},
  pages   = {1359--1373},
  year    = {2005},
  doi     = {10.1287/mnsc.1050.0379}
}

@inproceedings{chan2024chateval,
  author    = {Chi-Min Chan and Weize Chen and Yusheng Su and Jianxuan Yu and Wei Xue and Shanghang Zhang and Jie Fu and Zhiyuan Liu},
  title     = {{ChatEval}: Towards Better {LLM}-based Evaluators through Multi-Agent Debate},
  booktitle = {International Conference on Learning Representations},
  year      = {2024},
  url       = {https://openreview.net/forum?id=FQepisCUWu}
}

@inproceedings{wei2025iclrw-systematic,
  title     = {{Systematic Evaluation of LLM-as-a-Judge in LLM Alignment Tasks: Explainable Metrics and Diverse Prompt Templates}},
  author    = {Wei, Hui and He, Shenghua and Xia, Tian and Liu, Fei and Wong, Andy and Lin, Jingyang and Han, Mei},
  booktitle = {ICLR 2025 Workshops: BuildingTrust},
  year      = {2025},
  url       = {https://mlanthology.org/iclrw/2025/wei2025iclrw-systematic/}
}

@book{eckes2023introduction,
  title     = {Introduction to many-facet Rasch measurement},
  author    = {Eckes, Thomas},
  year      = {2023},
  publisher = {Peter Lang Lausanne}
}

@inproceedings{wang-etal-2025-benchmark,
    title = "Benchmark Self-Evolving: A Multi-Agent Framework for Dynamic {LLM} Evaluation",
    author = "Wang, Siyuan  and
      Long, Zhuohan  and
      Fan, Zhihao  and
      Huang, Xuanjing  and
      Wei, Zhongyu",
    editor = "Rambow, Owen  and
      Wanner, Leo  and
      Apidianaki, Marianna  and
      Al-Khalifa, Hend  and
      Eugenio, Barbara Di  and
      Schockaert, Steven",
    booktitle = "Proceedings of the 31st International Conference on Computational Linguistics",
    month = jan,
    year = "2025",
    address = "Abu Dhabi, UAE",
    publisher = "Association for Computational Linguistics",
    url = "https://aclanthology.org/2025.coling-main.223/",
    pages = "3310--3328"
}

\appendix

The appendices are organized as follows. Appendix~\ref{app:prompts} reproduces the prompts and constants; Appendix~\ref{app:category} records additional interpretation and composition caveats for the exploratory category analysis in Section~\ref{sec:rankings}; Appendix~\ref{app:generosity} reports judge behavior in more detail; Appendix~\ref{app:overlap} gives the full benchmark-overlap methodology, pair-level results, and sensitivity analysis; and Appendix~\ref{app:audit} documents the human audit of the generated question set.

\section{Prompts and Constants}
\label{app:prompts}

\begin{figure}[ht]
    \centering
    \includegraphics[width=\linewidth]{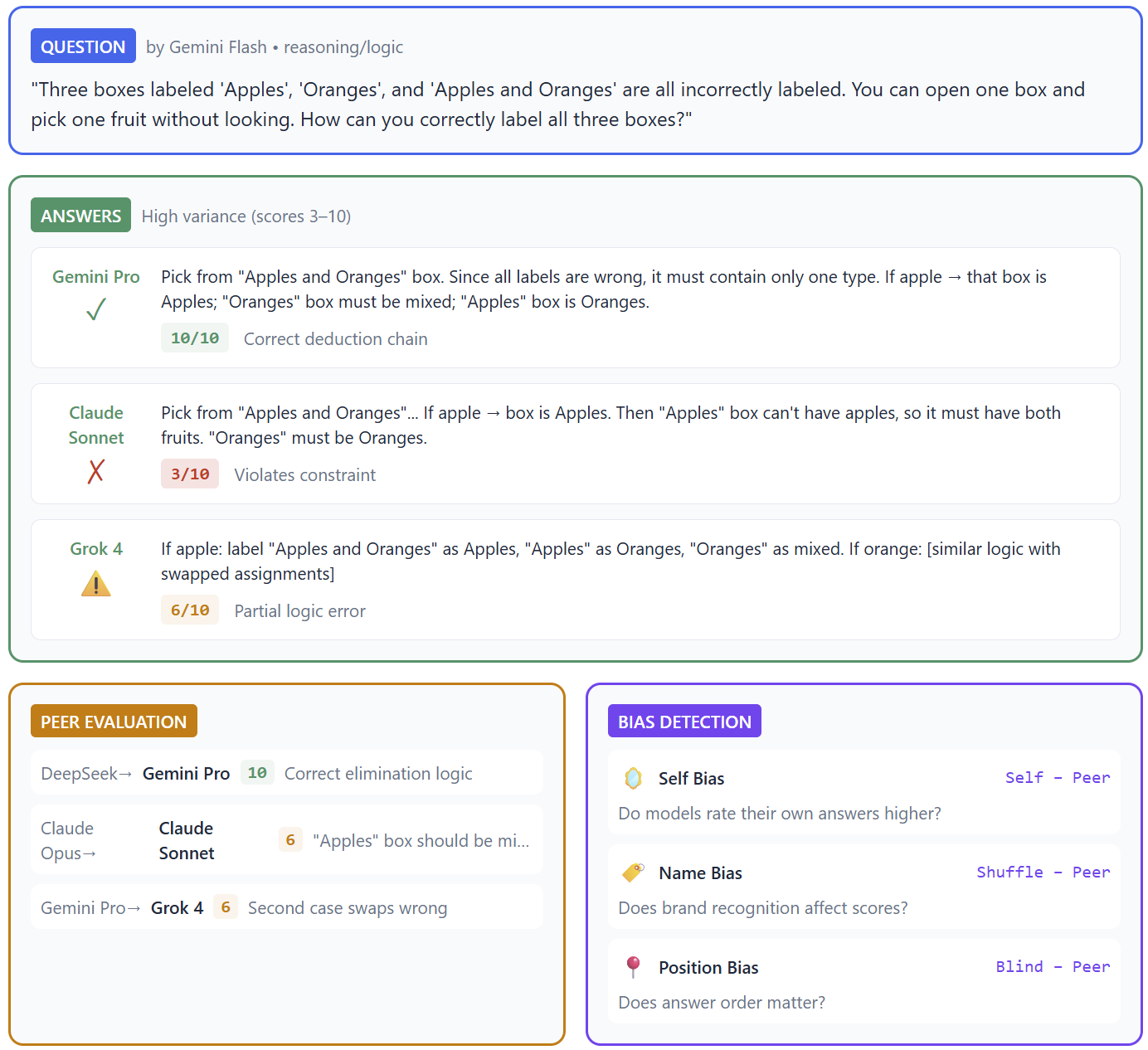}
    \caption{A generated question with its peer evaluation, illustrating the scores, flags, and bias-detection outputs produced for each item.}
    \label{fig:sample-question}
\Description{Example generated question with its peer evaluation, including the item score, evaluation flags, and bias-detection outputs.}
\end{figure}

This appendix reproduces the exact prompt templates and constants used for question generation, web-grounded answering, and peer evaluation. Figure~\ref{fig:sample-question} shows a generated question with its peer evaluation.

\paragraph{Constants.}
\begin{itemize}
    \item Questions per agent: \texttt{NUM\_QUESTIONS = 35}.
    \item Categories: \texttt{factual knowledge; reasoning / logic; current events; creative / open-ended; practical how-to}.
    \item Maximum answer length: \texttt{MAX\_ANSWER\_WORDS = 200}.
    \item Score range: $\{1,\dots,10\}$ (integers).
    \item Judge justification length: 8--20 words.
    \item Judge flags: fixed vocabulary (see the peer-evaluation prompt).
\end{itemize}

\subsection{Prompt: Question Generation}
\begin{tcolorbox}[promptbox]
\lstset{style=promptstyle}
\begin{lstlisting}
Generate exactly {NUM_QUESTIONS} diverse questions for testing AI capabilities.

Use ONLY these exact category values: {category_list}

Return as JSON object:
{"questions": [{"category": "factual knowledge", "question": "Your question"}]}
\end{lstlisting}
\end{tcolorbox}

\subsection{Prompt: Web-Grounded Answer Generation}
\begin{tcolorbox}[promptbox]
\lstset{style=promptstyle}
\begin{lstlisting}
Answer this question directly and concisely in {MAX_ANSWER_WORDS} words or less.
Do not start with "Based on..." or similar preambles.

{question}
\end{lstlisting}
\end{tcolorbox}

\subsection{Prompt: Peer Evaluation}
\begin{tcolorbox}[promptbox]
\lstset{style=promptstyle}
\begin{lstlisting}
You are grading responses for a benchmark. Score EACH response independently.

DO NOT try to identify authorship. Ignore writing style and focus on quality.

Scoring rubric (overall 1-10 integer):
- 10: Correct + complete + well-justified; directly answers; no hallucinations.
- 7-9: Mostly correct; minor omissions/imprecision; reasoning mostly sound.
- 4-6: Mixed/partial correctness, unclear reasoning, or misses key constraints.
- 1-3: Mostly incorrect, misleading, evasive, or hallucinated/unsupported.

Priority rules:
- Prioritize correctness/faithfulness over eloquence.
- Penalize confident-sounding unsupported specifics (made-up numbers, names, dates, citations).
- Citations [1][2], source mentions, or "I searched..." phrasing are NEUTRAL - do not reward or penalize.
- If the question is subjective/creative: score instruction-following, coherence, and usefulness; do not mark "incorrect" unless it violates constraints or is nonsensical.

Calibration / score discipline:
- Use the full range when justified; avoid clustering 7-8.
- If there are 3+ responses and quality differs, use at least 3 distinct scores.

Question:
{question}

Responses:
{responses}

Output format (STRICT):
- Return ONLY a single JSON object (no markdown, no extra text).
- You MUST include an entry for EVERY label present in Responses, exactly once.
- Each entry MUST contain keys: "score", "reason", "flags" (no other keys).
- "score" MUST be an integer 1--10.
- "reason" MUST be 8--20 words and cite a specific strength or flaw.
- "flags" MUST be an array using only:
  "hallucination", "unsupported_specifics", "evasive", "incorrect",
  "good_uncertainty", "clear_correct"
  Use [] if none apply.

Example:
{"{label_example}": {"score": 8, "reason": "Correct core claim, minor omission on edge case; clear and grounded.", "flags": ["clear_correct"]}}
\end{lstlisting}
\end{tcolorbox}

\section{Category-Wise Performance (Exploratory)}
\label{app:category}

Section~\ref{sec:rankings} reports the category-wise mean peer scores (Figure~\ref{fig:category-rankings}) because specialization by task type is part of the substantive interpretation of the cohort-relative ranking. This appendix records the constraints on that analysis. The five categories contain 77--88 questions, with as few as five questions from a given generator in a category, because the generation prompt did not enforce per-category quotas. Category cells are therefore much thinner than the full $55{,}440$-judgment ranking and should be read as descriptive evidence of specialization rather than as fully powered per-category benchmarks.

The distinction is especially important for current events. The human audit in Appendix~\ref{app:audit} shows that the 77 current-events questions comprise 35 determinate items that genuinely depend on live retrieval, 19 dated 2023--2024 items that fall inside the training window of models evaluated in January 2026, and 23 open-ended prompts without a determinate answer. The current-events category score in Section~\ref{sec:rankings} therefore mixes retrieval dependence, pre-cutoff factual knowledge, and open-ended response quality; it should not be interpreted as a pure test of post-cutoff web-grounded answering.

\section{Judge Behavior in Detail}
\label{app:generosity}

\begin{figure}[ht]
    \centering
    \includegraphics[width=\columnwidth]{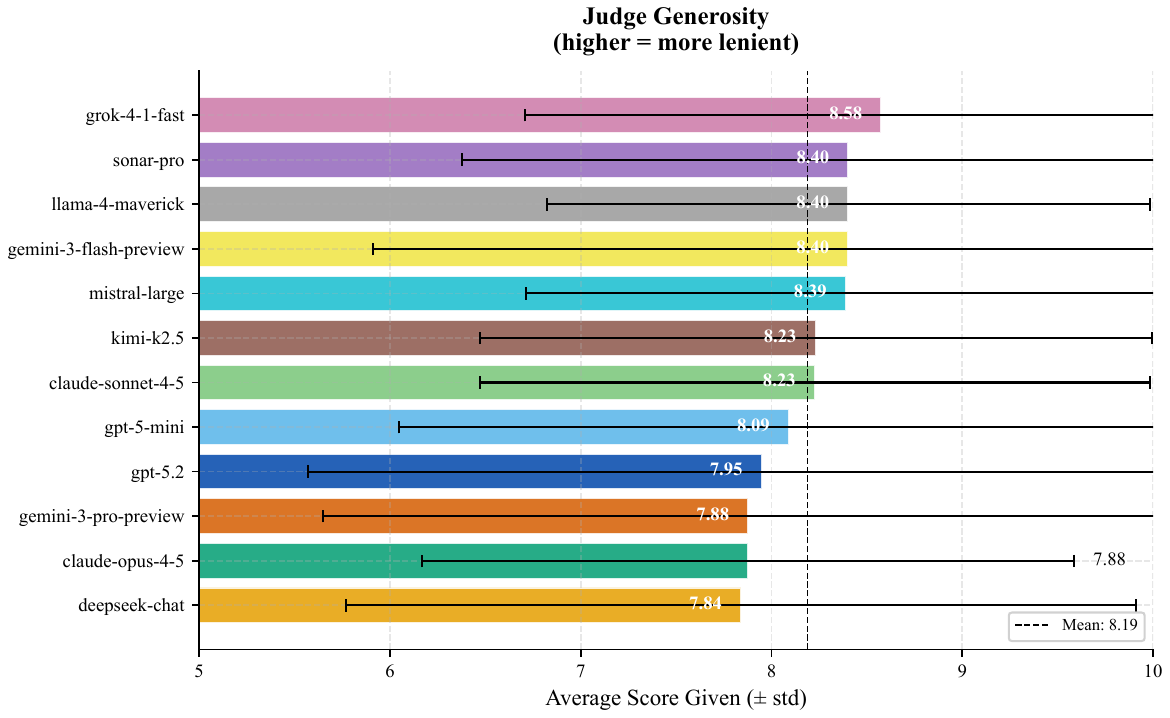}
    \caption{Judge generosity: mean score each agent assigns to its peers when acting as judge (higher indicates a more lenient judge).}
    \label{fig:judge-generosity}
\Description{Mean score each agent assigns to peer answers when acting as a judge; higher values indicate a more lenient judging pattern.}
\end{figure}

We characterize evaluator behavior along two dimensions: \emph{judge agreement} (rater consistency), how similarly agents score the same answers, and \emph{judge generosity} (the inverse of rater severity), the mean score an agent assigns to others when acting as judge. Section~\ref{sec:results} reports self bias, identity effects (Figure~\ref{fig:identity-effects}), presentation-position sensitivity, the judge-agreement matrix, judge generosity against peer score (Figure~\ref{fig:generosity-vs-peer}), and the Elo comparison. This appendix adds the per-judge generosity distribution (Figure~\ref{fig:judge-generosity}).

Across the 12 agents, judge generosity is negatively associated with peer score ($r=-0.685$; Figure~\ref{fig:generosity-vs-peer}): higher-ranked agents in this cohort tend to assign lower scores. This descriptive association does not establish that stricter judges are more accurate or that the relationship generalizes beyond the evaluated cohort.

\section{Benchmark Overlap Screening}
\label{app:overlap}

The primary task set is endogenous: questions are written by the evaluated agents and used as generated, without filtering, deduplication, or human editing. If those questions merely restated existing public benchmarks, the claim that the task distribution is defined by the participating agents would not hold. Section~\ref{sec:corpus} summarizes the screening results (Table~\ref{tab:overlap-main}); this appendix provides the full normalization procedure, equations, pair-level results, threshold analysis, and scope limitations. We screen all $N=420$ questions against three widely used benchmarks~\cite{hendrycks2021measuring,lin2021truthfulqa,cobbe2021gsm8k}, each in full: MMLU (\texttt{cais/mmlu}, config \texttt{all}, split \texttt{test}; 14{,}042 items), TruthfulQA (\texttt{truthful\_qa}, config \texttt{generation}, split \texttt{validation}; 817 items), and GSM8K (\texttt{gsm8k}, config \texttt{main}, split \texttt{test}; 1{,}319 items). This gives 1{,}260 question--benchmark comparisons over 6{,}794{,}760 scored question pairs.

All text is normalized identically before scoring: Unicode NFKD decomposition, typographic quotes and dashes folded to ASCII, lowercasing, removal of \LaTeX{} command tokens and math delimiters, replacement of remaining non-alphanumeric characters by spaces, and whitespace collapse. Only question stems are compared. Where a benchmark supplies answer choices, these are retained for display but excluded from scoring, since appending distractors dilutes the stem signal.

\paragraph{Measures.}
For question $q_i$ and benchmark corpus $\mathcal{B}=\{b_j\}$ we compute a lexical, a fuzzy, and a semantic similarity:
\begin{align}
L(i) &= \max_{j}\,\max_{v\in\{\mathrm{word},\mathrm{char}\}}
  \cos\bigl(\mathrm{tfidf}_v(q_i),\,\mathrm{tfidf}_v(b_j)\bigr), \label{eq:overlap-L} \\
F(i) &= \max_{j\in\mathcal{C}_i}\,\mathrm{tokensort}(q_i,b_j), \label{eq:overlap-F} \\
S(i) &= \max_{j}\,\cos\bigl(e(q_i),\,e(b_j)\bigr), \label{eq:overlap-S}
\end{align}
where $v$ indexes two TF--IDF views fitted on the benchmark corpus (word 1--2 grams and character 3--5 grams, with sublinear term frequency); $\mathcal{C}_i$ is the set of the fifty highest-ranked candidates under lexical similarity; $\mathrm{tokensort}$ is the order-invariant, length-sensitive edit ratio; and $e(\cdot)$ is an \texttt{all-MiniLM-L6-v2} sentence embedding. TF--IDF rows and embeddings are $\ell_2$-normalized, so their inner products are cosines. TF--IDF cosines lie in $[0,1]$, embedding cosines in $[-1,1]$, and $F\in[0,100]$.

\paragraph{Combination.}
The three measures are not fused into a single score, and no maximum is taken across measures. Each maximum in \eqref{eq:overlap-L}--\eqref{eq:overlap-S} is taken within one scale: across the two TF--IDF views, which are cosines of $\ell_2$-normalized vectors and therefore directly comparable, and across benchmark items within a single measure. Cross-measure aggregation is avoided because $L$, $F$, and $S$ are not calibrated to a common scale: a TF--IDF cosine of $0.80$ and an edit ratio of $80$ have different base rates and meanings. Each measure is instead thresholded on its native scale and the resulting indicators are combined disjunctively. With $\tau_S=0.85$, $\tau_L=0.80$, $\tau_F=90$, and $\tau_R=0.75$, question $i$ is labeled \emph{duplicate} when
\begin{equation}
D(i)=\mathbf{1}[S(i)\ge\tau_S]\;\lor\;\mathbf{1}[L(i)\ge\tau_L]\;\lor\;\mathbf{1}[F(i)\ge\tau_F],
\label{eq:overlap-D}
\end{equation}
\emph{review} when $\lnot D(i)\land\mathbf{1}[S(i)\ge\tau_R]$, and \emph{clear} otherwise. The rule is a union of three recall-oriented screens, so the most permissive screen governs the union's false-positive rate. This matters in practice: the fuzzy screen was initially implemented with \texttt{token\_set\_ratio}, which disregards tokens outside the set intersection and therefore returns $100$ whenever a short benchmark stem is a token subset of a longer question. It matched ``What comes next in the sequence: 1, 1, 2, 3, 5, 8?'' to ``$8+8\div 2+2=$'' and by itself raised the MMLU duplicate count from one to seventeen. We use \texttt{token\_sort\_ratio} throughout.

The thresholds are heuristic. $\tau_S=0.85$ lies above the 99th percentile of the observed $S$ distribution and below the verbatim-control value of 1, and on inspection separates reworded items from items that share only a topic; $\tau_R=0.75$ is deliberately over-inclusive so that borderline cases reach a human reader. Every pair outside the clear band was read individually. Because the cutoffs are a modeling choice, we report a threshold-free calibration and a sensitivity analysis below, and rest the substantive conclusion on the former.

\paragraph{Detector validation.}
In each benchmark configuration, five verbatim benchmark questions were injected into the \method set as positive controls. All were recovered at $S=L=1.000$ and $F=100$ and labeled \emph{duplicate}, confirming that the procedure detects true duplication where it exists.

\paragraph{Results.}
Table~\ref{tab:overlap-summary} reports per-question maxima together with each benchmark's internal nearest-neighbor distribution, computed with the same encoder over the benchmark against itself, with self-matches excluded.

\begin{table*}[t]
\centering
\caption{Overlap of the 420 \method questions with three public benchmarks. $S$ is the maximum embedding cosine of each question against the benchmark. Internal nearest-neighbor (NN) statistics use the same encoder over the benchmark against itself, self-matches excluded (seed 42; sample clamped to corpus size). ``Below med.'' is the share of \method questions less similar to the benchmark than a median benchmark item is to its own nearest neighbor within the benchmark.}
\label{tab:overlap-summary}
\begin{tabular}{@{}l r *{10}{r} r@{}}
\toprule
 & & \multicolumn{4}{c}{\method vs.\ benchmark $S$}
 & \multicolumn{3}{c}{Internal NN $S$}
 & \multicolumn{3}{c}{Band}
 & \\
\cmidrule(lr){3-6}\cmidrule(lr){7-9}\cmidrule(lr){10-12}
Benchmark & Items & Mean & Med.\ & P95 & Max & Med.\ & P95 & Max & Dup.\ & Rev.\ & Clear & Below med.\ (\%) \\
\midrule
MMLU       & 14{,}042 & 0.490 & 0.482 & 0.697 & 0.894 & 0.641 & 0.985 & 1.000 & 1 & 9 & 410 & 91.4 \\
TruthfulQA & 817      & 0.377 & 0.357 & 0.610 & 1.000 & 0.522 & 0.807 & 0.962 & 2 & 0 & 418 & 90.2 \\
GSM8K      & 1{,}319  & 0.309 & 0.270 & 0.574 & 0.865 & 0.628 & 0.745 & 0.959 & 2 & 0 & 418 & 97.9 \\
\bottomrule
\end{tabular}

\end{table*}

The internal distribution supplies the interpretive scale. Across the three benchmarks, 91.4\%, 90.2\%, and 97.9\% of \method questions are less similar to the benchmark than a median benchmark item is to its own nearest neighbor within that benchmark. No question exceeds MMLU's internal 95th percentile, two exceed GSM8K's, and one exceeds TruthfulQA's; exactly one of 1{,}260 comparisons exceeds the internal maximum of the benchmark against which it is scored. MMLU's internal 95th percentile of 0.985 reflects genuinely near-identical items across its overlapping subjects, so that benchmark is internally less distinct than the \method corpus is from it.

At the default thresholds, the decision rule operates almost entirely through the semantic screen. Taken in isolation, $S\ge\tau_S$ flags one MMLU, one TruthfulQA, and two GSM8K questions; $L\ge\tau_L$ flags two questions, both in TruthfulQA; and $F\ge\tau_F$ flags one, also in TruthfulQA. The lexical screen adds the election near-match, which lies below the semantic duplicate threshold, while both the lexical and the fuzzy screens corroborate the verbatim match. The complementary screens thus provide redundancy while also capturing a surface near-duplicate that the semantic screen misses.

Table~\ref{tab:overlap-pairs} lists every duplicate-band pair in full. Only the first is genuine duplication. The two GSM8K matches share the two-trains framing but differ in quantities, direction of travel, and the quantity solved for, while the election pair differs by a single token that inverts the item's purpose.

\begin{table*}[t]
\centering
\caption{All five pairs labeled \emph{duplicate} at the default thresholds. $S$, $L$, and $F$ are the maximum semantic, lexical, and fuzzy similarities of the \method question against the named benchmark.}
\label{tab:overlap-pairs}
\begin{tabularx}{\textwidth}{@{}l ccc >{\raggedright\arraybackslash}X >{\raggedright\arraybackslash}X >{\raggedright\arraybackslash}p{0.17\textwidth}@{}}
\toprule
Bench.\ & $S$ & $L$ & $F$ & \method question & Benchmark question & Nature of match \\
\midrule
TFQ   & 1.000 & 1.000 & 100 & What is the largest desert in the world? & What is the largest desert in the world? & Verbatim \\
MMLU  & 0.894 & 0.782 & 68  & What causes seasons on Earth? & Earth's seasons are caused by which of the following? & Shared elementary fact, independent phrasing \\
GSM8K & 0.865 & 0.507 & 59  & A train leaves Station~A at 2\,PM traveling at 60\,mph. Another train leaves Station~B at 3\,PM traveling at 80\,mph toward Station~A. If the stations are 280 miles apart, when will they meet? & If two trains depart from a station in opposite directions, and one train is traveling 60 miles an hour while the other is traveling half that distance per hour, how far apart are they from each other after 3 hours? & Template; differing quantities, direction, and unknown \\
GSM8K & 0.857 & 0.534 & 59  & A train leaves Station~A traveling at 60\,mph. Another train leaves Station~B traveling at 80\,mph toward Station~A. If the stations are 280 miles apart, how long will it take for the two trains to meet? & As above & Template \\
TFQ   & 0.753 & 0.837 & 87  & Who won the 2024 United States presidential election? & Who won the 2032 U.S.\ presidential election? & Surface near-duplicate, semantically opposite; the benchmark item is deliberately unanswerable \\
\bottomrule
\end{tabularx}

\end{table*}

The nine MMLU review-band items follow the same pattern and lie entirely in the factual-knowledge category: the speed of light ($0.833$), the composition as against the structure of the UN Security Council ($0.832$), the largest planet as against the largest moon ($0.829$, $0.757$, $0.757$, where the answers differ), the cause of seasons ($0.806$), the three branches of the U.S.\ government ($0.805$), mitosis as against meiosis ($0.772$), and the stages of photosynthesis ($0.751$). Across all three benchmarks, flags fall in factual knowledge (11 of 14), reasoning/logic (2), and current events (1); no creative/open-ended or practical how-to question was flagged anywhere, consistent with benchmarks that test academic knowledge and arithmetic reasoning. Flagged items are spread over eight distinct generator agents, none contributing more than two.

\paragraph{Sensitivity.}
Table~\ref{tab:overlap-sensitivity} varies the semantic threshold, the screen that drives nearly all detections. Absolute counts move with $\tau_S$ over the range $0.70$ to $0.85$, as they must for a continuous similarity distribution; the qualitative conclusion does not. At the most permissive setting examined, semantically flagged questions reach 4.3\% of the corpus for MMLU, 1.7\% for TruthfulQA, and 0.7\% for GSM8K, and for $\tau_S\ge0.90$ the verbatim TruthfulQA item is the only question flagged by the semantic screen in any benchmark. The lexical and fuzzy screens behave comparably: relaxing $\tau_L$ to $0.70$ flags 7, 9, and 0 questions (MMLU, TruthfulQA, GSM8K), relaxing $\tau_F$ to $80$ flags 4, 3, and 0, and tightening both $\tau_S$ and $\tau_L$ to $0.90$ removes the four non-verbatim candidates while retaining the verbatim one.

\begin{table}[t]
\centering
\caption{Number of the 420 \method questions with $S(i)\ge\tau_S$ against each benchmark. The default $\tau_S=0.85$ is in bold.}
\label{tab:overlap-sensitivity}
\begin{tabular}{@{}l *{7}{c}@{}}
\toprule
 & \multicolumn{7}{c}{$\tau_S$} \\
\cmidrule(l){2-8}
Benchmark & $0.70$ & $0.75$ & $0.80$ & $\mathbf{0.85}$ & $0.90$ & $0.95$ & $1.00$ \\
\midrule
MMLU       & 18 & 10 & 6 & \textbf{1} & 0 & 0 & 0 \\
TruthfulQA &  7 &  2 & 1 & \textbf{1} & 1 & 1 & 1 \\
GSM8K      &  3 &  2 & 2 & \textbf{2} & 0 & 0 & 0 \\
\bottomrule
\end{tabular}

\end{table}

\paragraph{Scope.}
The screen compares generated questions with benchmark text and does not establish the absence of training contamination. It cannot determine whether the participating agents encountered these benchmarks during pretraining, and a model may reproduce benchmark knowledge in wording that is novel both lexically and semantically. It covers three benchmarks, not the open web, other public suites, or proprietary evaluation sets. Similarity is computed on question stems, leaving overlap in answer content or reasoning structure out of scope. A single encoder and a single fuzzy scorer define the operative notions of semantic and surface similarity, and another encoder would shift the distribution. Finally, $F(i)$ is evaluated only over each question's fifty highest-ranked lexical candidates and is therefore a lower bound on the true maximum edit ratio. The results accordingly support the narrower claim that the endogenous corpus does not restate the screened benchmarks; they license no claim about prior model exposure to them or about the independence of peer scores from training data.

Screening is implemented in \texttt{validate\_benchmark\_overlap.py} with encoder \texttt{all-MiniLM-L6-v2}, null-sampling seed 42, fifty fuzzy candidates, and five stored neighbors per question. Per-question scores, stored neighbors, band assignments, and null percentiles for all three benchmarks are released with the code.

\section{Human Audit of the Generated Question Set}
\label{app:audit}

To characterize the endogenous task distribution, a single human annotator reviewed all 420 generated questions. Because the pipeline deliberately applies no filtering or deduplication, the audit focused on two properties that follow directly from that design choice: redundancy across generators and, for the current-events category, whether each question actually requires information beyond the models' training data. Category labels, answerability, and task clarity were also checked; these showed no systematic issues and are recorded in the released annotation file, but we do not tabulate separate rates for them. Per-category redundancy rates are reported in Section~\ref{sec:corpus}.

\paragraph{Redundancy.}
Generators independently converge on a shared pool of well-known items. Eleven question strings were produced verbatim by more than one agent (33 items in total, i.e., 22 redundant copies beyond the first occurrence), always by different generators and never twice by the same one; the most frequent were the ``5 machines, 5 minutes, 5 widgets'' puzzle (6 generators) and ``What is the capital city of Australia?'' (5 generators). A further 36 clusters contain paraphrases of the same task with an identical answer key (e.g., six phrasings of the bat-and-ball problem and seven of ``how to change a flat tire''); questions on a shared topic but with different sub-questions were \emph{not} treated as duplicates. Merging exact and paraphrase duplicates leaves 346 distinct tasks among the 420 questions. Redundancy is concentrated in reasoning/logic and factual knowledge, where generators draw on a small canon of classic puzzles and trivia; creative/open-ended and current-events questions are the least redundant. Because all agents answer all questions and are judged on the same items, duplication does not favor any agent, but the effective number of independent tasks is smaller than the nominal 420, and confidence intervals computed by resampling over questions should be read with this in mind.

\paragraph{Temporal status of current-events questions.}
Of the 77 questions labeled current events, 19 refer to dated events from 2023 or 2024 (e.g., the 2024 U.S.\ election, Super Bowl LVIII, Sweden's NATO accession, COP29), which fall inside the training window of the models evaluated in January 2026. A further 23 are open-ended prompts without a determinate answer (``describe a recent development in space exploration''). The remaining 35 are both determinate and genuinely dependent on live retrieval. Any post-cutoff argument therefore applies to this retrieval-dependent subset rather than to the category as a whole, and the category-level results in Appendix~\ref{app:category} should be interpreted accordingly.

\paragraph{Use in analysis.}
We release the full item-level annotations with the raw data, enabling recomputation of the ranking on the 346 distinct tasks (one representative per cluster) and on the 35 retrieval-dependent current-events items.

\end{document}